\documentclass{article}
\PassOptionsToPackage{table}{xcolor}
\usepackage[T1]{fontenc}
\usepackage[utf8]{inputenc}
\usepackage{microtype}
\usepackage{graphicx}
\usepackage{booktabs}
\usepackage{hyperref}

\usepackage[preprint]{icml2026}
\usepackage{enumitem}
\usepackage{needspace}
\setlist[itemize,enumerate,description]{topsep=3pt,partopsep=0pt,parsep=0pt,itemsep=0pt}
\usepackage{amsmath}
\usepackage{amssymb}
\usepackage{multirow}
\usepackage{xcolor}
\usepackage{colortbl}
\usepackage{pgfplots}
\pgfplotsset{compat=1.18}
\usetikzlibrary{arrows.meta,positioning}

\newcommand{\fitbox}[1]{\setbox0\hbox{#1}\leavevmode\ifdim\wd0>\columnwidth\resizebox{\columnwidth}{!}{\box0}\else\box0\fi}
\newcommand{\fitboxw}[1]{\setbox0\hbox{#1}\leavevmode\ifdim\wd0>\textwidth\resizebox{\textwidth}{!}{\box0}\else\box0\fi}

\newcommand{\ldm}{LDM}
\newcommand{\dds}{DDS}
\newcommand{\sprd}{SPRD}

\icmltitlerunning{Large Decision Models}

\begin{document}

\twocolumn[
  \icmltitle{Large Decision Models: Repurposing Large Language Models\\as Single-Pass Decision Makers}
  \icmlsetsymbol{equal}{*}
  \begin{icmlauthorlist}
    \icmlauthor{Ziyi Chen}{uiuc,equal}
    \icmlauthor{Yan Zhang}{nus,equal}
    \icmlauthor{Jianhui Wei}{zju,equal}
    \icmlauthor{Jianing Yin}{upenn}
    \icmlauthor{Daoan Zhang}{rochester}
  \end{icmlauthorlist}
  \icmlaffiliation{uiuc}{University of Illinois Urbana-Champaign}
  \icmlaffiliation{nus}{National University of Singapore}
  \icmlaffiliation{zju}{Zhejiang University}
  \icmlaffiliation{upenn}{University of Pennsylvania}
  \icmlaffiliation{rochester}{University of Rochester}
  \begin{center}
  \small
  \textsuperscript{1}University of Illinois Urbana-Champaign\quad
  \textsuperscript{2}National University of Singapore\\
  \textsuperscript{3}Zhejiang University\quad
  \textsuperscript{4}University of Pennsylvania\quad
  \textsuperscript{5}University of Rochester\\
  \textsuperscript{*}Equal contribution.
  \end{center}
  \icmlkeywords{Large Decision Models, schema transfer, distillation, evaluation}
  \vskip 0.3in
]

\raggedbottom
\makeatletter\global\icml@noticeprintedtrue\makeatother
\hypersetup{pdftitle={Large Decision Models: Repurposing Large Language Models as Single-Pass Decision Makers},pdfauthor={Ziyi Chen, Yan Zhang, Jianhui Wei, Jianing Yin, Daoan Zhang}}

\begin{abstract}
Large language models make decisions through parallel prefill and sequential decoding.
Although effective, this approach incurs decoding costs and does not explicitly expose decision probabilities.
Existing task-specific classifiers require per-task training and offer limited generalization to unseen decision schemas.
We propose Large Decision Models (\ldm{}s), repurposing language models to return explicit probabilities over labels specified at run time using one prefill forward pass.
We first adapt models to this interface through Diverse Decision Supervision (\dds{}), covering varied domains, rules and labels.
To further narrow the performance gap to a reasoning model, Self-Posed Reasoning Distillation (\sprd{}) adds tasks generated by a frozen teacher, filters them for answer consistency, and trains on label distributions obtained from repeated teacher reasoning.
We evaluate on DecisionVista, pairing new-schema tests with held-out human-labelled sources.
Across three base models, the recipe improves accuracy over a step-matched baseline by 1.8 and 1.5 percentage points, respectively, achieving a 182$\times$ speedup in latency over reasoning rollout on Gemma-4-31B.
Reasoning remains stronger on the new-schema test, motivating a confidence router for uncertain decisions.
On development sets, routing recovers a median 92\% of reasoning's accuracy advantage while generating 27--58\% of its tokens.
Code is available at \url{https://github.com/JianhuiWei7/LDM}.
\end{abstract}

\section{Introduction}
\label{sec:intro}
\begin{figure*}[t]
\centering
\includegraphics[width=\textwidth]{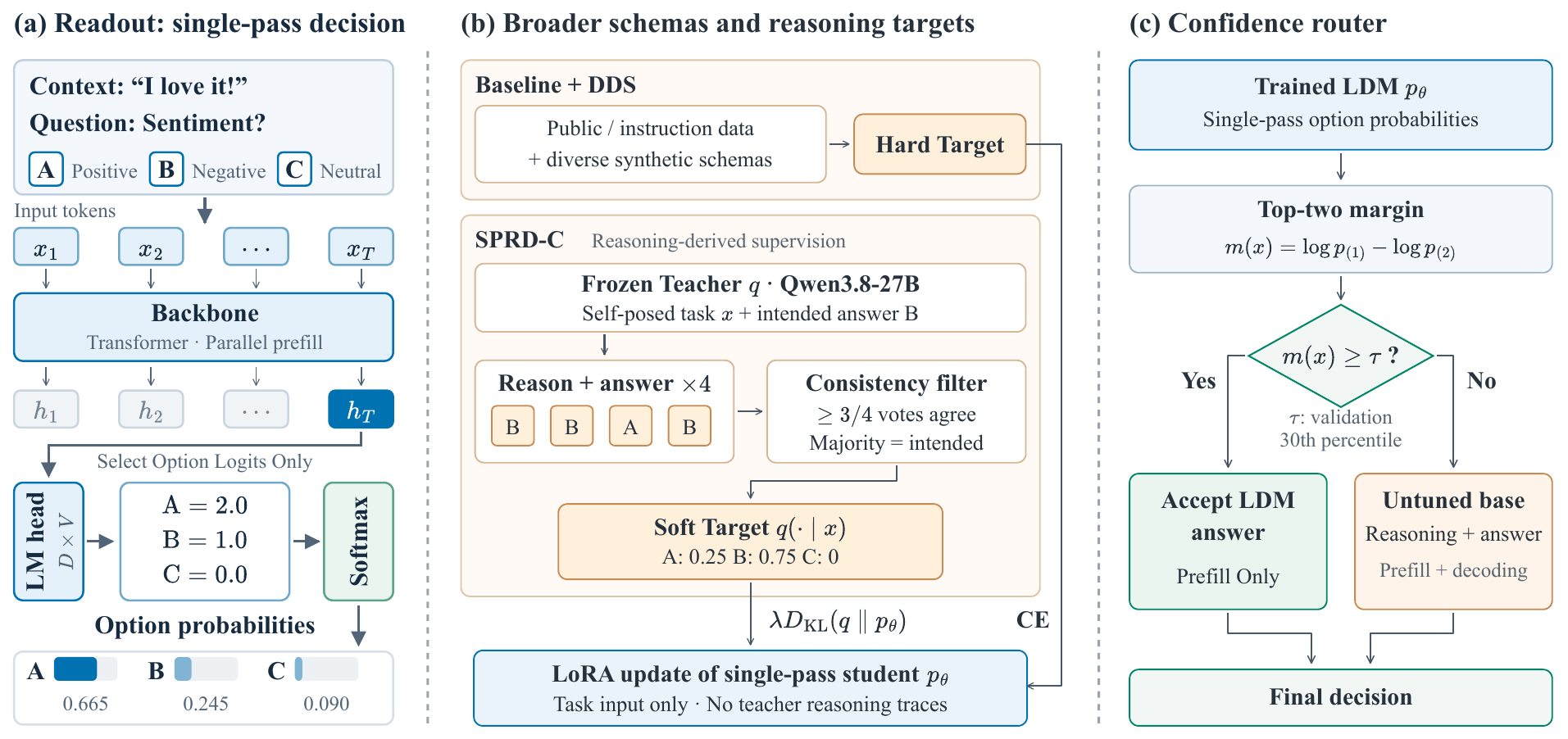}
\caption{Overview. (a) The LDM processes the task in one prefill, projects the last input hidden state through the LM head, and normalises only the allowed option logits. (b) Baseline and DDS examples supply hard-label cross-entropy supervision. The frozen Qwen3.8-27B teacher generates four reasoning-and-answer samples per self-posed task. SPRD-C filters for consistency and distils answer-vote distributions with $\lambda D_{\mathrm{KL}}(q\Vert p_\theta)$. (c) A margin router accepts confident LDM predictions and otherwise invokes the student's untuned base for one reasoning rollout.}
\label{fig:overview}
\end{figure*}

Many applications require a decision: route a support ticket, judge whether a claim is supported, or rate an incident's severity.
Large Language Models (LLMs) acquire broad knowledge through pretraining and generalize across diverse tasks.
However, generating decisions through autoregressive decoding (\emph{rollout}) incurs latency and computational costs and does not explicitly expose decision probabilities, limiting their suitability for tasks that require fast responses and confidence-aware choices.
Conventional BERT-based classifiers offer efficient predictions, but task-specific fine-tuning ties their output heads to predefined labels~\citep{devlin2019bert}.
To combine classification efficiency with the generality of pretrained language models, we study \textbf{Large Decision Models} (\ldm{}s), which map a context, a question and a label set specified at run time to a distribution over the labels in a single forward pass---an operation we call \emph{readout} (\S\ref{sec:ldm}; Fig.~\ref{fig:overview}).
Existing systems such as Jev, Nimble, OpenJev and Laya offer this interface~\citep{jev,nimble,openjev,laya}.

To adapt LLM rollout to LDM readout, we use \emph{decision tuning}: given a context, a question and candidate labels, we fine-tune the model with LoRA to predict the correct label by applying cross-entropy to the normalized option-token logits at the final input position, without generating any sequence.
After training, we evaluate \emph{schema transfer}: accuracy on decision schemas, label sets and domains absent from fine-tuning, alongside calibration and single-pass cost.
Schema transfer is feasible because pretraining already places most of the decision in the representation that a single forward pass reads. Without any decision training, the readout of an untuned 27B model reaches 97\% of its tuned accuracy, and 1/16 of the tuning data recovers 88\% of the tuning gain at 9B (\S\ref{sec:diagnosis}). Decision tuning is therefore an \emph{adaptation} of how that representation is read out, not the acquisition of new task knowledge.

Our diagnosis identifies two opportunities for improving transfer (\S\ref{sec:diagnosis}).
First, decision tuning improves source accuracy more reliably than new-schema accuracy, motivating supervision across broader schemas.
Second, reasoning rollout reveals capabilities that the single-pass readout does not fully express: on two new-schema validation sets, the same untuned 27B model is 7.8--9.3 points more accurate when allowed to reason.
This rollout--readout gap motivates two complementary uses of reasoning: distillation during training and selective computation at inference.

Our training recipe builds on these two opportunities (\S\ref{sec:method}).
\textbf{Diverse Decision Supervision (\dds{})} uses synthetic decision schemas to promote generalization beyond familiar tasks and output patterns.
Varying domains, rules and label meanings, alongside public and instruction data, encourages the readout to interpret each new schema.
\textbf{\mbox{Self-Posed Reasoning} Distillation (\sprd{})} aims to bring readout performance closer to that of the base model using reasoning rollout.
It distils a reasoning teacher's answer distributions on new self-posed tasks into the readout, making those decisions a training target while retaining single-pass inference.

At deployment, a \textbf{confidence router} combines fast readout with selective reasoning rollout.
It accepts confident readout answers and sends uncertain decisions to the base model for reasoning rollout, using a threshold fixed on validation data (\S\ref{sec:routing}).

We evaluate \dds{} and \sprd{} separately and together across three base models from two families, with three seeds each (\S\ref{sec:results}).
We introduce \textbf{DecisionVista}, a benchmark suite pairing new decision schemas (\textbf{DecisionVista-Synth}) with human-labelled tasks from eight sources excluded from fine-tuning and method design (\textbf{DecisionVista-Human}).
Our protocol separates model development from pre-registered testing on these two sets (\S\ref{sec:evaluation}).
Averaged across bases, the combined recipe improves accuracy over a step-matched baseline by 1.8 points on DecisionVista-Synth and 1.5 on DecisionVista-Human.
Against a baseline selected on validation data, it gains 0.8 points on DecisionVista-Human and 2.1 on the public typed-decisions benchmark.
Our 9B model outperforms Nimble-9B on the same base across transfer evaluations and Nimble-13.
On Gemma-4-31B served on two H100 GPUs, tuned readout reduces median latency from 7.38\,s to 40.6\,ms, a 182$\times$ speedup over reasoning rollout (Table~\ref{tab:speed}).
On validation sets, routing recovers a median 92\% of the rollout--readout gap using 27--58\% of full reasoning's generated tokens (\S\ref{sec:router}).

\par\begingroup
\setlength{\parskip}{0pt}
\interlinepenalty=10000
\Needspace{4\baselineskip}
\noindent Our contributions are:\par\nopagebreak
\begin{enumerate}[topsep=0pt]
\item \textbf{\ldm{} formulation:} single-pass decision models centered on generalization to unseen schemas.
\item \textbf{\dds{} and \sprd{}:} a training recipe combining synthetic schema diversity with distillation of a reasoning teacher's decisions.
\item \textbf{DecisionVista:} synthetic and human-labelled benchmarks, with a protocol separating model development from pre-registered testing.
\item \textbf{Confidence routing:} a deployment strategy combining fast readout with selective reasoning rollout.
\item \textbf{Statistically validated gains:} our recipe improves accuracy over the step-matched baseline by 1.8 points on DecisionVista-Synth (95\% CI [0.6, 3.1]) and 1.5 on DecisionVista-Human (95\% CI [0.9, 2.2]). Tuned readout achieves a 182$\times$ median-latency speedup over reasoning rollout on Gemma-4-31B.
\end{enumerate}
\endgroup

\section{Large Decision Models}
\label{sec:ldm}

A \emph{Large Decision Model} (\ldm{}) maps a context $c$ and a decision schema specified at run time to a probability distribution over the schema's allowed outcomes in a single forward pass.
A schema consists of a question or decision rule $q$ and a finite label set $\mathcal{Y}$ with natural-language descriptions:
\begin{equation}
p_\theta(\cdot \mid c, q, \mathcal{Y}) \in \Delta(\mathcal{Y}),
\label{eq:ldm}
\end{equation}
where $\Delta(\mathcal{Y})$ is the set of probability distributions over $\mathcal{Y}$; the predicted outcome is $\arg\max_{y\in\mathcal{Y}} p_\theta(y\mid c,q,\mathcal{Y})$.
Schemas may express categorical choices, binary judgements or ordinal ratings, with label meanings and any ordering supplied at run time.
This allows the same model to address new decisions without a task-specific output head or further training.

A pretrained model can use this interface zero-shot. Decision tuning aligns the model with this interface by adapting autoregressive rollout to single-pass readout.
Jev~\citep{jev} serves this interface as a hosted service; Nimble-9B~\citep{nimble} and OpenJev-27B~\citep{openjev} release LLM weights for it, and Laya~\citep{laya} a 0.3--0.4B encoder with a trained decision head.
An \ldm{} should be judged first on \emph{schema transfer}, accuracy on decision schemas, label sets and domains absent from fine-tuning, and then on calibration and cost.

\section{Evaluating schema transfer}
\label{sec:evaluation}

\begin{table}[t]
\centering\small
\caption{Evaluation sets and primary uses (DV: DecisionVista; Val.: validation suite). Validation sets guide development; the validation suite supports selection and calibration. Test sets were sealed and first read under pre-registered analyses after the models were frozen; later analyses are marked post hoc. No model in the primary study trains on these sets. $\geq$2 of 3 LLMs: kept iff at least two of three blind non-author labellers agree with the author; $^\ast$DV-Pilot was sealed under its earlier rule (Appendix~\ref{app:chronology}). Grey rows appear only in the appendix.}
\label{tab:evalsets}
\setlength{\tabcolsep}{2pt}
\fitbox{%
\begin{tabular}{@{}lrrll@{}}
\toprule
Set & Items & Tasks & Labels & Primary use \\
\midrule
Nimble benchmark & 3,880 & 13 & human & Public benchmark \\
Typed-decisions (test) & 2,000 & 4 & LLM teacher & Public benchmark \\
Validation suite & 1,256 & 7 & human & Validation \\
Transfer suite & 2,955 & 8 & human & Validation \\
DV-Dev-A/B & 1,177 & 24 & $\geq$2 of 3 LLMs & Validation \\
\textbf{DV-Synth} & 1,505 & 35 & $\geq$2 of 3 LLMs & \textbf{Test} \\
\textbf{DV-Human} & 2,299 & 8 & human & \textbf{Test} \\
\textcolor{gray}{DV-Stress} & \textcolor{gray}{1,134} & \textcolor{gray}{24} & \textcolor{gray}{$\geq$2 of 3 LLMs} & \textcolor{gray}{Validation} \\
\textcolor{gray}{DV-Pilot} & \textcolor{gray}{1,311} & \textcolor{gray}{29} & \textcolor{gray}{3 LLMs agree$^\ast$} & \textcolor{gray}{Test} \\
\bottomrule
\end{tabular}}
\end{table}

Testing schema transfer requires both comparison with existing systems and evidence that gains extend beyond the tasks used to develop the recipe.
We therefore separate public comparisons, model selection and exploratory diagnosis from independent confirmation (Table~\ref{tab:evalsets}).

\paragraph{Public benchmarks and development sets.}
Public benchmarks anchor comparisons with released decision models: Nimble's full 13-subset benchmark (\emph{Nimble-13})~\citep{nimble} and the typed-decisions test set reported for Jev and Laya (\emph{Typed-Decisions})~\citep{laya}; four workflows, labels from an LLM teacher.
Typed-decisions contains 2,000 decisions across 400 cases (five per case); its authors report 73.5\% repeatability of the teacher labels.
No model in the primary study trains on their sources.
Table~\ref{tab:released} also reports Kev's transfer-v4 test (\emph{Kev-Transfer}; 656 items from sources new to Kev, including MMLU and QNLI), read once.

The human-labelled \emph{validation suite} (Val.) supports model selection, temperature calibration and routing thresholds.
Other validation sets probe complementary forms of transfer: the human-labelled \emph{transfer suite} tests generalization across sources, while synthetic DV-Dev-A/B test newly constructed decision schemas.
These sets guide recipe development, so gains on them require confirmation on untouched sets.
The human-suite sources are excluded from fine-tuning, and synthetic evaluation families are screened against training schemas.
DV-Dev-A/B contain rule-based decisions written in contrastive groups by gpt-5.5 and kept when gpt-6-astra agreed blind.
Source inventories and the earlier DV-Stress and DV-Pilot stages are in Appendices~\ref{app:repro} and~\ref{app:chronology}.

\paragraph{DecisionVista tests.}
The two primary sets were constructed after the primary models were frozen, then sealed and first read under pre-registered analyses; no design decision used them.
\textbf{DecisionVista-Synth} tests newly constructed schemas: 35 families over 18 decision operators, including capacity allocation, Pareto choice and counterfactual intervention.
Four model families author the tasks; the three non-authors label each item blind, and at least two must agree with the author (Appendix~\ref{app:fresh5}).
Families are screened against earlier evaluation families and training schemas; we claim screened non-overlap, not freedom from pretraining exposure.
Its labels are model agreement, not human judgement; a human audit protocol is released.
We also evaluate without the author that verified synthetic \dds{} labels (Appendix~\ref{app:provenance}).
\textbf{DecisionVista-Human} complements this test with 2,299 human-labelled items from eight sources absent from all training mixtures, earlier evaluation sets and Super-NaturalInstructions tasks (Appendix~\ref{app:humanseal}).
Here the held-out units are sources; related task types can appear in training.
Tables abbreviate the pair to DV-Synth and DV-Human.

\paragraph{Measurement and protocol.}
The primary statistics are mean accuracy across task families on DecisionVista-Synth and across sources on DecisionVista-Human, averaged equally across bases.
Paired bootstrap intervals resample task families or items within each human-labelled source; a two-level sensitivity analysis also resamples training seeds (Appendix~\ref{app:stats}).
Nimble scores average its 13 subset accuracies; typed-decisions intervals resample its 400 cases.
Calibration uses negative log-likelihood (NLL), Brier score and expected calibration error (ECE; 15 bins).
Appendix~\ref{app:chronology} records which sets informed each choice; analyses specified after the sealed reads are marked \emph{post hoc}.
Pre-registrations, hashes, per-item predictions and code are in the supplementary material.

\section{Diagnosing schema transfer}
\label{sec:diagnosis}

Our exploratory diagnosis asks why decision tuning transfers unevenly and how much accuracy the single-pass readout leaves unused.
Two observations motivate broader schema supervision and reasoning-derived targets.

\paragraph{The decision is largely already in the representation.}
Across Qwen3.5 sizes from 0.8B to 27B, the untuned single-pass readout reaches 76--97\% of the accuracy of the same model after decision tuning, the tuning gain shrinks from 15.8 to 2.3 points as the base grows, and 1/16 of the tuning data already recovers most of it (Appendix~\ref{app:adapt}).
Pretraining thus supplies the representation; tuning adapts how it is read. The adaptation is light but must reach the backbone: LoRA updates under 0.5\% of the parameters, whereas training only the option-letter head on frozen features recovers almost none of the gain (Appendix~\ref{app:head}).

\paragraph{Source gains do not ensure schema transfer.}
We rebuilt Nimble's recipe and varied scale, data, objective and capacity (Appendix~\ref{app:recipes}).
At 27B, training on 20k records from 36 public datasets raises validation accuracy from 79.3 to 81.8; adding the public benchmark's training sources gives 89.1 on its eight choice and yes/no subsets, versus OpenJev's 84.1.
Yet neither recipe consistently improves new-schema accuracy: the public-data recipe scores 88.1/87.5 on DV-Dev-A/B and the source-augmented recipe 87.8/88.0, against the untuned model's 89.1/87.5.
Transfer depends on scale and evaluation set---baseline tuning does help at 9B---so this is a motivation to broaden schema coverage, not a claim that supervised tuning cannot transfer.

\paragraph{The rollout--readout gap.}
The same untuned 27B model is more accurate when allowed to reason (majority of four samples): 96.9 vs.\ 89.1 on DV-Dev-A, 96.8 vs.\ 87.5 on DV-Dev-B and 85.1 vs.\ 79.3 on the validation suite (Table~\ref{tab:devgap}).
Where the two modes disagree on DV-Dev-B, reasoning is right 90\% of the time.
A separate single-sample analysis finds the largest gains on exact multi-step computation (dates and numeric comparisons: 56--66\% readout, 98--100\% reasoning across the three bases), while reasoning can lose on NLI (Appendix~\ref{app:reasoning}).
Reasoning costs hundreds of generated tokens per decision.
These observations motivate \dds{} to broaden the schemas encountered during training and \sprd{} to train the readout toward a reasoning teacher's decisions, paying for teacher computation offline.
Confidence routing retains access to reasoning for uncertain decisions at deployment.

\section{Training \ldm{}s for schema transfer}
\label{sec:method}

We train the readout to interpret unfamiliar decision schemas and to approximate decisions available from a reasoning teacher.
\dds{} broadens schema coverage through synthetic decisions alongside public and instruction data; \sprd{} supplies reasoning-derived answer distributions on self-posed tasks.
At deployment, a confidence router can send uncertain decisions to the student's untuned base using reasoning rollout.

\paragraph{Implementation.}
For this study, we implement the \ldm{} interface by normalising the logits of single-token option identifiers:
\begin{equation}
p_\theta(y \mid c, q, \mathcal{Y}) = \operatorname*{softmax}_{y \in \mathcal{Y}} \, z_\theta\big(\ell_y \mid x(c, q, \mathcal{Y})\big),
\label{eq:readout}
\end{equation}
where $x(\cdot)$ serialises the context, question and options, $\ell_y$ is the letter assigned to outcome $y$, and $z_\theta(\ell_y\mid x)$ is its logit at the final input position.
We support \emph{choice} (2--8 named options), \emph{yes/no}, and \emph{ordinal score} (3--6 ordered levels); these bounds are implementation choices, not part of the \ldm{} definition.
Each instruction-tuned base trains a LoRA adapter~\citep{hu2022lora} on this readout, with option order re-shuffled at every draw.
The baseline training data contain 22.4k public and contrastive records (Appendix~\ref{app:spec}).

\subsection{Diverse Decision Supervision (\dds{})}
\label{sec:dds}
Learning a common answer format does not by itself ensure that the readout can apply unfamiliar decision rules.
\dds{} uses synthetic task construction to vary domains, rules and label meanings, aiming to encourage schema interpretation beyond the original training tasks.
Public datasets and categorical instruction tasks~\citep{wang2022supernatural} provide complementary coverage, while the readout objective stays fixed.

Each run uses the same 31.7k-record sample from the 63.4k pool: additional public sources, Super-NaturalInstructions tasks, and synthetic decisions built from a domain--decision grid or grounded in real documents (Appendix~\ref{app:dds}).
Synthetic labels require blind verification, and synthetic schemas are audited against evaluation families.
The pool excludes benchmark-source data and is 13-gram de-contaminated against evaluation items available at construction; the later DecisionVista sets are checked against the frozen mixtures (Appendices~\ref{app:fresh5},~\ref{app:humanseal}).

\subsection{Self-Posed Reasoning Distillation (\sprd{})}
\label{sec:sprd}
\sprd{} turns the rollout--readout gap into supervision: the readout learns decisions reached with additional computation, without learning to generate the reasoning trace.
Qwen3.8-27B is the frozen task author and reasoning teacher for all primary students; only the 27B student learns from its own untuned base.

\paragraph{Self-posed tasks and decision targets.}
The teacher proposes schemas across domains, reasoning skills and answer formats, then writes contrastive instances for schemas that pass an evaluation-family audit: 16,040 items from 696 schemas (Appendix~\ref{app:repro}).
It answers each item four times using reasoning rollout.
The target is $q(y\mid x)=n_y/m$, where $n_y$ counts votes for $y$ and $m$ is the number of parseable samples.
Requiring the top vote share over all four samples to be at least 0.75 retains 14,829 items.
The consistency-filtered variant, \textbf{\sprd{}-C}, further keeps the 10,331 items whose reasoning majority agrees with the label intended at item construction.
This checks consistency between two computations of the same model, not correctness.
The student receives answer distributions, never reasoning traces.

\paragraph{Joint training.}
Baseline data and \dds{} records supply supervised labels; pool records supply teacher distributions.
We shuffle the enabled data streams together and minimise their mean per-record loss:
\begin{equation}
\ell_\theta(r) = \begin{cases}
-\log p_\theta(y^\star_r\mid x_r), & r\in\mathcal{D}_{\text{sup}},\\[2pt]
\lambda\,D_{\mathrm{KL}}\big(q_r\,\|\,p_\theta(\cdot\mid x_r)\big), & r\in\mathcal{D}_{\sprd{}},
\end{cases}
\label{eq:loss}
\end{equation}
where $\mathcal{D}_{\text{sup}}$ contains baseline data plus \dds{} when enabled, and $\lambda=3$.
Inference remains one forward pass.

\subsection{Confidence routing to reasoning}
\label{sec:routing}
The router accepts confident readout decisions and sends uncertain items to that student's untuned base using reasoning rollout (one sample).
Its score is the top-two margin $m(x)=\log p_\theta(y_{(1)}\mid x)-\log p_\theta(y_{(2)}\mid x)$, equal to the corresponding logit gap in Eq.~\ref{eq:readout}.
If $m(x)<\tau$, the reasoning answer is used; otherwise the readout's answer stands at zero generated tokens.
We fix $\tau$ at the $b$-quantile of margins on the validation suite, with $b=30\%$ unless stated, before reading other sets.
This sets validation coverage; the fraction routed on an unseen set can differ.
The router requires no additional training.

\section{Results}
\label{sec:results}

\begin{table*}[!t]
\centering\small
\caption{Accuracy gains over the step-matched baseline (points; 95\% bootstrap intervals; bold: interval above zero). Shading marks the combined recipe. Its DV-Synth and DV-Human means across bases are the pre-registered primary tests; other contrasts are secondary or descriptive.}
\label{tab:contrast}
\setlength{\tabcolsep}{2.8pt}
\fitboxw{%
\begin{tabular}{llcccc}
\toprule
 & & Development & \multicolumn{3}{c}{Test (used in no design decision)} \\
\cmidrule(lr){3-3}\cmidrule(lr){4-6}
 & & & Public & \multicolumn{2}{c}{Sealed, pre-registered} \\
\cmidrule(lr){4-4}\cmidrule(lr){5-6}
Base & vs.\ step-matched baseline & Transfer + DV-Dev-A/B & Typed-Decisions & DV-Synth & DV-Human \\
\midrule
Qwen3.5-9B & +\dds{} & \textbf{+3.2} {\scriptsize[1.7, 4.7]} & \textbf{+2.4} {\scriptsize[1.0, 3.8]} & \textbf{+1.6} {\scriptsize[0.07, 3.2]} & \textbf{+2.2} {\scriptsize[1.2, 3.2]} \\
 & +\sprd{}-C & \textbf{+3.2} {\scriptsize[1.6, 4.8]} & +1.3 {\scriptsize[$-$0.2, 2.8]} & \textbf{+2.1} {\scriptsize[0.02, 4.0]} & \textbf{+2.1} {\scriptsize[1.1, 3.1]} \\
\rowcolor{gray!15}  & +\dds{} + \sprd{}-C & \textbf{+3.9} {\scriptsize[2.4, 5.5]} & \textbf{+2.5} {\scriptsize[0.9, 4.2]} & \textbf{+2.3} {\scriptsize[0.6, 3.9]} & \textbf{+2.1} {\scriptsize[1.1, 3.1]} \\
\midrule
Qwen3.8-27B & +\dds{} & \textbf{+1.6} {\scriptsize[0.5, 2.7]} & \textbf{+1.7} {\scriptsize[0.5, 2.9]} & +1.2 {\scriptsize[$-$0.9, 3.2]} & +1.0 {\scriptsize[$-$0.04, 2.1]} \\
 & +\sprd{}-C & \textbf{+1.9} {\scriptsize[0.9, 3.0]} & \textbf{+3.1} {\scriptsize[1.8, 4.4]} & +1.6 {\scriptsize[$-$1.0, 4.6]} & +0.9 {\scriptsize[$-$0.05, 1.9]} \\
\rowcolor{gray!15}  & +\dds{} + \sprd{}-C & \textbf{+2.3} {\scriptsize[1.4, 3.4]} & \textbf{+3.0} {\scriptsize[1.7, 4.3]} & +2.4 {\scriptsize[$-$0.3, 5.3]} & \textbf{+1.9} {\scriptsize[0.8, 3.0]} \\
\midrule
Gemma-4-31B & +\dds{} & \textbf{+1.8} {\scriptsize[0.8, 2.9]} & \textbf{+1.5} {\scriptsize[0.5, 2.6]} & +1.4 {\scriptsize[$-$0.3, 3.1]} & +0.1 {\scriptsize[$-$1.0, 1.1]} \\
 & +\sprd{}-C & \textbf{+1.9} {\scriptsize[1.0, 2.9]} & +0.9 {\scriptsize[$-$0.3, 2.2]} & $-$0.2 {\scriptsize[$-$1.9, 1.4]} & $-$0.1 {\scriptsize[$-$1.0, 0.9]} \\
\rowcolor{gray!15}  & +\dds{} + \sprd{}-C & \textbf{+2.5} {\scriptsize[1.5, 3.6]} & +1.3 {\scriptsize[$-$0.2, 2.7]} & +0.7 {\scriptsize[$-$0.7, 2.2]} & +0.6 {\scriptsize[$-$0.4, 1.5]} \\
\midrule
Mean across bases & +\dds{} & +2.2 & \textbf{+1.9} {\scriptsize[1.1, 2.6]} & \textbf{+1.4} {\scriptsize[0.4, 2.4]} & \textbf{+1.1} {\scriptsize[0.4, 1.8]} \\
 & +\sprd{}-C & +2.3 & \textbf{+1.8} {\scriptsize[0.8, 2.7]} & +1.1 {\scriptsize[$-$0.1, 2.5]} & \textbf{+1.0} {\scriptsize[0.4, 1.6]} \\
\rowcolor{gray!15}  & +\dds{} + \sprd{}-C & +2.9 & \textbf{+2.3} {\scriptsize[1.3, 3.2]} & \textbf{+1.8} {\scriptsize[0.6, 3.1]} & \textbf{+1.5} {\scriptsize[0.9, 2.2]} \\
\midrule
\multicolumn{2}{l}{Both vs.\ validation-selected baseline, mean (post hoc)} & -- & \textbf{+2.1} {\scriptsize[1.2, 3.0]} & +0.8 {\scriptsize[$-$0.1, 1.7]} & \textbf{+0.8} {\scriptsize[0.2, 1.4]} \\
\bottomrule
\end{tabular}}
\end{table*}

We first test transfer on the pre-registered evaluations, then examine component effects, compare released models and evaluate confidence routing.
We compare baseline training alone, with \dds{}, with \sprd{}-C, and with both on Qwen3.5-9B, Qwen3.8-27B and Gemma-4-31B-it, using the fixed Qwen teacher from \S\ref{sec:sprd}.
The primary comparator, \textbf{step-matched baseline}, trains for 4,032 steps against the full recipe's 4,031 at batch size 16 (2.9 passes over baseline data).
We report three seeds per variant; the pre-registered test estimates use seeds 17/29/41 at 9B and 17/29 at 27B/31B, excluding later seed-41 runs (Appendix~\ref{app:spec}).
The validation column averages accuracy on DV-Dev-A, DV-Dev-B and the transfer suite.

\paragraph{Primary transfer results.}
Both pre-registered primary tests succeed (Table~\ref{tab:contrast}).
The combined recipe gains \textbf{+1.8 points} over the step-matched baseline on DecisionVista-Synth (95\% CI [0.6, 3.1]; sign-flip $p{=}0.009$) and \textbf{+1.5} on DecisionVista-Human ([0.9, 2.2]).
The gains are larger for Qwen: +2.3/+2.1 on the two sets at 9B and +2.4/+1.9 at 27B, compared with +0.7/+0.6 for Gemma, whose intervals include zero.
Neither set is near ceiling (step-matched baseline: 64.5--80.0 on DecisionVista-Synth, 72.8--79.6 on DecisionVista-Human; Appendix Table~\ref{tab:fresh5}).

\paragraph{Dependence on the baseline comparator.}
Matching steps controls training compute but can overtrain on the smaller baseline dataset.
In a \emph{post-hoc} comparison, we select the baseline on the validation suite from one-pass or step-matched training and full or 0.65-scaled updates, selecting the recipe only over full or 0.65-scaled updates on the same suite.
The gains become +0.8 on DecisionVista-Synth ([$-$0.1, 1.7]), +0.8 on DecisionVista-Human ([0.2, 1.4]) and +2.1 on typed-decisions ([1.2, 3.0]; Appendix Table~\ref{tab:devsel}).
A well-selected baseline recovers most of the synthetic-schema advantage; the human-labelled and typed-decisions gains remain above zero.

\paragraph{Component effects and validation results.}
The original single-component variants train fewer steps (\dds{}: 3,386; \sprd{}-C: 2,050), so their positive mean effects over the step-matched baseline are descriptive.
On DecisionVista-Human, the combination gains +0.4 [0.0, 0.9] over \dds{} and +0.5 [0.1, 1.0] over \sprd{}-C.
On DecisionVista-Synth, the pre-registered combination-versus-\dds{} test is not significant (+0.4 [$-$0.3, 1.1]); the ordered component-testing chain stops there (Appendix Table~\ref{tab:secondary}).
\emph{Post-hoc} single-component runs at 4,031 steps still beat step-matched baseline on DecisionVista-Human at 9B and 27B; at 27B both have similar mean gains to the combination on DecisionVista-Synth, while only \sprd{}-C does so on typed-decisions (Appendix Table~\ref{tab:stepcomp}).

On the transfer suite and DV-Dev-A/B, the combination gains +3.9, +2.3 and +2.5 across the three bases, with all 27 per-seed differences positive; against the one-pass baseline the gains are +4.0, +1.6 and +1.4.
These sets informed design.
The pooled test gains are 52--62\% of the validation gain (+2.9).
Primary intervals remain positive under seed resampling, alternative novelty units, the author exclusion and inclusion of later seeds; dropping any one human-labelled source retains a positive mean gain (Appendix~\ref{app:moved}).

\begin{table*}[tp]
\centering\small
\caption{Decision accuracy (\%; three-seed means for our tuned models; step-matched baselines). Bold: best per block, excluding reasoning references ($^\circ$). Shaded: our recipe. $^\star$Confidence router with the same setting on every base: the 30\% least confident items of the validation suite define the threshold, and routed items are answered by the untuned base using reasoning rollout (one sample); generated-token costs are in \S\ref{sec:router}. $^\circ$Untuned base using reasoning rollout on every item. $^\P$Trained on BoolQ and MultiNLI; $^\ddagger$trained on typed-decisions (TD); $^\S$published results, not rerun. Benchmark and scoring details: \S\ref{sec:evaluation} and \S\ref{sec:results}.}
\label{tab:released}
\setlength{\tabcolsep}{3.6pt}
\fitboxw{%
\begin{tabular}{lrrrrr}
\toprule
 & \multicolumn{3}{c}{Public benchmarks (test)} & \multicolumn{2}{c}{Ours: development / sealed test} \\
\cmidrule(lr){2-4}\cmidrule(lr){5-6}
Model & Nimble-13 & Kev-Transfer & Typed-Decisions & Transfer suite & DV-Human \\
 & {\color{black!55}\scriptsize Nimble; 13 tasks} & {\color{black!55}\scriptsize Kev; new sources} & {\color{black!55}\scriptsize Laya/Jev; workflows} & {\color{black!55}\scriptsize dev.; 8 MC sources} & {\color{black!55}\scriptsize sealed; 8 sources} \\
\midrule
\multicolumn{6}{l}{\emph{Released decision models}} \\
Jev 1.13.0$^\S$ & \textbf{76.0} & -- & 72.7 & -- & -- \\
Laya-TD$^\S$ & -- & -- & \textbf{76.6} & -- & -- \\
Laya-TD, 0.4B$^\ddagger$ & 59.8 & 67.1 & 76.2 & 35.5 & 51.6 \\
Laya-en (0.4B) & 57.2 & 66.5 & 37.4 & 35.0 & 51.0 \\
Kev-4B$^\P$ & 74.3 & \textbf{83.8} & 66.7 & \textbf{65.6} & \textbf{69.2} \\
\midrule
\multicolumn{6}{l}{\emph{Qwen3.5-9B base}} \\
Untuned & 73.7 & 77.0 & 63.5 & 66.2 & 69.4 \\
\textcolor{gray}{Reasoning rollout}$^\circ$ & \textcolor{gray}{76.1} & -- & \textcolor{gray}{72.2} & \textcolor{gray}{75.0} & \textcolor{gray}{76.3} \\
Nimble-9B & 74.8 & 82.5 & 65.0 & 67.5 & 72.5 \\
Kev-9B$^\P$ & 75.3 & 85.2 & 71.4 & 65.9 & 72.6 \\
Baseline & 74.1 & 84.9 & 69.8 & 69.7 & 72.8 \\
\rowcolor{gray!15} \dds{} + \sprd{}-C & 76.2 & \textbf{85.8} & 72.3 & 71.9 & 74.9 \\
\rowcolor{gray!15} + router$^\star$ & \textbf{76.7} & -- & \textbf{73.2} & \textbf{76.2} & \textbf{76.4} \\
\midrule
\multicolumn{6}{l}{\emph{Qwen3.8-27B base}} \\
Untuned & 74.4 & 85.5 & 72.0 & 75.4 & 73.2 \\
\textcolor{gray}{Reasoning rollout}$^\circ$ & \textcolor{gray}{72.8} & -- & \textcolor{gray}{75.2} & \textcolor{gray}{80.6} & \textcolor{gray}{76.0} \\
OpenJev-27B & 77.7 & 88.3 & 73.0 & 76.5 & 76.5 \\
Kev-27B$^\P$ & \textbf{77.8} & 89.5 & 75.9 & 73.5 & 75.5 \\
Baseline & 75.2 & \textbf{90.3} & 71.7 & 76.2 & 75.2 \\
\rowcolor{gray!15} \dds{} + \sprd{}-C & 75.7 & \textbf{90.2} & 74.7 & 77.5 & \textbf{77.0} \\
\rowcolor{gray!15} + router$^\star$ & 74.0 & -- & \textbf{76.4} & \textbf{80.6} & 76.4 \\
\midrule
\multicolumn{6}{l}{\emph{Gemma-4-31B base}} \\
Untuned & \textbf{77.8} & 88.6 & 75.0 & 77.8 & 79.2 \\
\textcolor{gray}{Reasoning rollout}$^\circ$ & \textcolor{gray}{77.0} & -- & \textcolor{gray}{74.8} & \textcolor{gray}{82.2} & \textcolor{gray}{80.9} \\
Baseline & 75.3 & \textbf{90.3} & 74.1 & 77.6 & 79.5 \\
\rowcolor{gray!15} \dds{} + \sprd{}-C & 76.3 & 90.2 & 75.4 & 79.0 & 80.2 \\
\rowcolor{gray!15} + router$^\star$ & 76.9 & -- & \textbf{75.6} & \textbf{81.7} & \textbf{80.6} \\
\bottomrule
\end{tabular}}
\end{table*}

\paragraph{Comparison with released decision models.}
Table~\ref{tab:released} separates model bases and discloses benchmark training.
All models we run receive identical inputs: our LLMs, Nimble-9B and OpenJev use Nimble's scorer; Laya and Kev use their released packages, including Kev's shipped temperature.

Without training on benchmark sources, our 9B recipe scores 76.2 ($\pm$0.6 over seeds) on Nimble's 13-subset benchmark versus Nimble-9B's 74.8 on the same base.
The lead is concentrated in Civil Comments and MultiNLI (per-subset scores and seed deviations in Appendix Table~\ref{tab:public13}).
The 9B recipe also improves on Nimble on the human-labelled transfer suite (71.9 vs.\ 67.5) and DecisionVista-Human (74.9 vs.\ 72.5); the 27B recipe scores 77.0 on DecisionVista-Human versus OpenJev's 76.5.
On typed-decisions, which no design decision used, the recipe gains +2.3 [1.3, 3.2] over the step-matched baseline across bases; Kev-27B remains stronger among 27B readouts, while Laya-TD uses this benchmark's training split.

Public-benchmark rankings depend on source exposure: adding the benchmark's training splits gives \dds{} 80.8 at 9B and 81.7 at 27B on Nimble's benchmark, with transfer-suite scores of 71.9 and 77.6.
These source-trained runs belong to a different comparison regime; Appendix Table~\ref{tab:indist} reports them alongside the released systems and their training-data disclosures.
The larger source-free recipes do not lead Nimble's benchmark (Table~\ref{tab:released}).

\begin{figure}[t]\centering\definecolor{c9}{HTML}{2A78D6}\definecolor{c27}{HTML}{EB6834}\definecolor{c31}{HTML}{1BAF7A}
\begin{tikzpicture}
\begin{axis}[width=0.93\columnwidth, height=0.6\columnwidth, xmin=-3, xmax=108, xtick={0,25,50,75,100},
  xlabel={Generated tokens (\% of reasoning on every item)}, ylabel={Accuracy (\%)}, grid=major, grid style={gray!18},
  tick label style={font=\scriptsize}, label style={font=\small}, axis x line*=bottom, axis y line*=left, legend style={font=\scriptsize, draw=none, fill=none, at={(0.5,-0.36)}, anchor=north, legend columns=2, /tikz/every even column/.append style={column sep=6pt}}, legend cell align=left]
\addplot[c9, thick, mark=*, mark size=1.4pt, forget plot] coordinates {(0.00,81.16) (9.04,83.41) (16.98,84.98) (29.79,86.86) (40.01,87.85) (48.15,88.06) (56.94,88.06) (70.90,87.98)};
\addplot[c9, dashed, forget plot] coordinates {(0.00,81.16) (100.00,87.44)};
\addplot[c9, only marks, mark=*, mark size=2.8pt, mark options={draw=white}, forget plot] coordinates {(40.01,87.85)};
\addplot[c9, only marks, mark=star, mark size=3.2pt, very thick, forget plot] coordinates {(100.00,87.44)};
\addplot[c27, thick, mark=square*, mark size=1.4pt, forget plot] coordinates {(0.00,86.11) (9.64,87.66) (17.96,88.72) (34.57,90.43) (44.74,90.92) (53.78,90.99) (62.43,91.17) (72.14,91.33)};
\addplot[c27, dashed, forget plot] coordinates {(0.00,86.11) (100.00,91.32)};
\addplot[c27, only marks, mark=square*, mark size=2.8pt, mark options={draw=white}, forget plot] coordinates {(44.74,90.92)};
\addplot[c27, only marks, mark=star, mark size=3.2pt, very thick, forget plot] coordinates {(100.00,91.32)};
\addplot[c31, thick, mark=diamond*, mark size=1.4pt, forget plot] coordinates {(0.00,86.91) (5.89,87.93) (12.95,88.98) (25.26,90.15) (35.98,90.56) (42.62,90.86) (49.44,90.88) (65.53,91.36)};
\addplot[c31, dashed, forget plot] coordinates {(0.00,86.91) (100.00,91.77)};
\addplot[c31, only marks, mark=diamond*, mark size=2.8pt, mark options={draw=white}, forget plot] coordinates {(35.98,90.56)};
\addplot[c31, only marks, mark=star, mark size=3.2pt, very thick, forget plot] coordinates {(100.00,91.77)};
\addlegendimage{c9, thick, mark=*, mark size=1.4pt}\addlegendentry{Qwen3.5-9B}
\addlegendimage{c27, thick, mark=square*, mark size=1.4pt}\addlegendentry{Qwen3.8-27B}
\addlegendimage{c31, thick, mark=diamond*, mark size=1.4pt}\addlegendentry{Gemma-4-31B}
\addlegendimage{gray, thick, mark=*, mark size=1.4pt}\addlegendentry{router, Val. budgets 5--70\%}
\addlegendimage{gray, only marks, mark=*, mark size=2.8pt}\addlegendentry{router, 30\% budget}
\addlegendimage{gray, only marks, mark=star, very thick, mark size=3.2pt}\addlegendentry{reasoning, every item}
\addlegendimage{gray, dashed}\addlegendentry{random routing}
\end{axis}
\end{tikzpicture}
\caption{Confidence routing: accuracy (mean of the transfer suite, DV-Dev-A and DV-Dev-B; seed means) against generated tokens as a share of reasoning on every item. Router thresholds are fixed on the validation suite at routing budgets of 5--70\%; large marker: 30\%. Dashed: random routing, expected accuracy at the same generated-token share. Star: reasoning on every item.}
\label{fig:router}\end{figure}
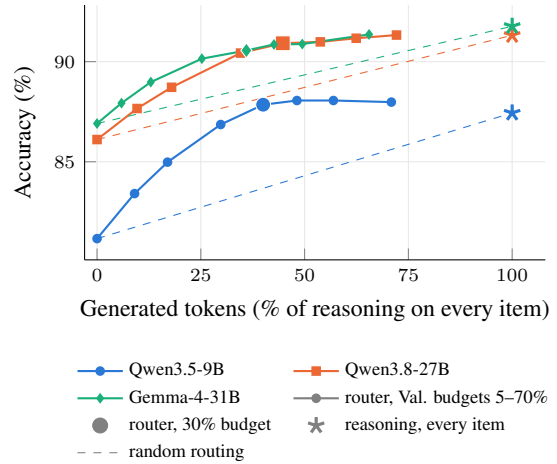

\begin{table}[t]
\centering\small
\caption{Gemma-4-31B serving efficiency. Speedup uses median latency relative to reasoning; GPU-hours per million requests are extrapolated.}
\label{tab:speed}
\setlength{\tabcolsep}{2pt}
\def\speedtablebody{\begin{tabular}{@{}lrrrr@{}}
\toprule
Mode & Median & Rate & Speedup & GPU-h \\
 & (ms) & (req./s) & ($\times$) & (/1M) \\
\midrule
\shortstack[l]{Reasoning\\rollout} & 7,380 & 0.787 & 1.0 & 706.2 \\
\multicolumn{1}{@{}>{\columncolor{gray!15}[0pt][\tabcolsep]}l}{\shortstack[l]{Tuned\\readout}} & \cellcolor{gray!15}40.6 & \cellcolor{gray!15}44.2 & \cellcolor{gray!15}181.8 & \multicolumn{1}{>{\columncolor{gray!15}[\tabcolsep][0pt]}r@{}}{12.6} \\
\bottomrule
\end{tabular}}
\setbox0=\hbox{\speedtablebody}
\setlength{\tabcolsep}{\dimexpr 2pt+(\columnwidth-\wd0)/8\relax}
\speedtablebody
\end{table}

\paragraph{Measured serving efficiency.}
On Gemma-4-31B served in BF16 on two H100 GPUs, tuned readout takes 40.6\,ms median latency versus 7.38\,s for reasoning rollout, with 56.2$\times$ higher throughput (Table~\ref{tab:speed}).
These measurements exclude confidence routing; checkpoint and measurement details are in Appendix~\ref{app:speed}.

\paragraph{Deployment through confidence routing.}\label{sec:router}
Table~\ref{tab:released} includes each untuned base using reasoning rollout with one sample as a reference; median generation costs range from 357 to 2,124 tokens per decision across models.
With the threshold set to route 30\% of the validation suite, the router recovers 66--140\% (median 92\%) of the rollout--readout gap on the transfer suite and DV-Dev-A/B while generating 27--58\% of reasoning's tokens.
Random routing of the same fraction of items recovers 20--35\% (Fig.~\ref{fig:router}; Appendix~\ref{app:router}).
On the transfer suite it matches reasoning ($+$1.2, 0.0, $-$0.5 points).
The test-set routing analyses are \emph{post hoc} and descriptive: DecisionVista-Human needs 52--80\% of reasoning's tokens, while DecisionVista-Synth gains 15--18 points over the readout but remains 2--7 below reasoning.
We count generated tokens only; routed items also require a second prefill, so end-to-end savings are smaller.

\section{Analysis}
\label{sec:analysis}

We examine what the supervision labels contribute, how well readout probabilities calibrate, and how the rollout--readout gap varies across bases.

\paragraph{What the labels contribute.}
\sprd{}-C adds new items, a filter and reasoning labels at once.
Holding the 10,331 filtered items and training steps fixed, replacing reasoning targets with the model's own single-pass labels costs 27B 1.3--2.4 points on the validation suite and DV-Dev-A/B, and 1.2--1.9 on DecisionVista-Human and typed-decisions (Appendix Tables~\ref{tab:samesubset},~\ref{tab:labelctl}).
The advantage over self-training therefore comes from the labels, conditional on the reasoning--intent filter that selected these items.
Hard and soft reasoning targets perform similarly.

\paragraph{Teacher identity and selection.}
Other capable labels (gpt-5.5 or the generator's intended answer) perform similarly on DecisionVista-Synth and DecisionVista-Human.
At 9B, \sprd{}-C exceeds an equal-sized random pool with generator labels by only +0.7 to +1.2 points at matched steps, with intervals touching zero.
The clearest advantage is over readout self-training; at 27B the model's own reasoning supplies the stronger labels without an additional labeller.
A post-hoc Gemma control also gives comparable results with its own reasoning labels in place of Qwen's (Appendix~\ref{app:gself}).
This is not a fixed-subset comparison: relabelling the same pool changes the retained items and adds about 2\% more training steps.
It does not support teacher family as the sole explanation for Gemma's smaller gains.

\paragraph{Calibration.}
Longer baseline training makes Qwen models overconfident (transfer-suite ECE 23.8 at 9B, 18.5 at 27B), while the recipes stay at 8--12.
One temperature fitted on the validation suite removes most of this gap and transfers to the test sets (Appendix Tables~\ref{tab:calib},~\ref{tab:sealedcalib}).
After scaling, every recipe retains a lower Brier score than the step-matched baseline on every base, and a lower NLL on both test sets; transfer-suite NLL at 27B is tied (0.71--0.73).
Kev's shipped temperature gives the lowest raw ECE but higher NLL and Brier than our Qwen recipes.
Gemma's raw readout remains strongly overconfident across training variants and requires calibration.

\paragraph{Mixture-of-experts bases.}
The smaller tested MoE bases leave larger rollout--readout gaps than the dense 27B model (Appendix~\ref{app:moe}).
Qwen3.5-122B-A10B scores 3.2 points higher using reasoning rollout on the validation suite but only ties its readout (79.8 vs.\ 79.3); 35B-A3B matches its reasoning while trailing its readout by two points.
The largest tested MoE, Qwen3.5-397B-A17B, instead has the same transfer-suite gap as the dense base (5.2 points) and a 2.6--2.8-point readout advantage.
On 35B-A3B, widening expert routing at inference or during tuning does not close the rollout--readout gap.
One possible explanation is that generating multiple reasoning tokens engages different experts before the answer, but these comparisons do not isolate that mechanism or establish a general scaling law.

\section{Related work}
\label{sec:related}

\paragraph{Decision interfaces and transfer.}
Label-aware classifiers such as TARS support unseen classes~\citep{halder2020tars}; class decisions can also be read from next-token scores~\citep{schick2021pet,zhao2021calibrate}.
Decision services and models~\citep{jev,openjev,nimble,laya,kev} standardise the typed interface we name and evaluate.
\dds{} applies instruction tuning and synthetic task construction~\citep{wei2022flan,sanh2022t0,wang2022supernatural,wang2023selfinstruct} to single-pass schema transfer.

\paragraph{Reasoning supervision.}
\sprd{} is closest to System-2 distillation~\citep{yu2024distilling} and implicit chain-of-thought~\citep{deng2023implicit,deng2024explicit}, but targets final-answer distributions at one readout position over run-time label sets.
Like on-policy (self-)distillation~\citep{agarwal2024onpolicy,zhao2026opsd}, it learns from stronger teacher behaviour; supervision here is offline and uses no student-generated trajectories.
Readout self-training~\citep{lee2013pseudo,xie2020noisy} supplies a label-source control.

\paragraph{Selective reasoning.}
The router combines selective prediction~\citep{geifman2017selective} with a same-base reasoning fallback, following the acceptance-and-escalation structure of LLM cascades~\citep{chen2023frugalgpt,ong2024routellm}.
Its margin threshold is fixed on the validation suite; probability calibration separately uses temperature scaling~\citep{guo2017calibration}.
Appendix~\ref{app:related} gives the fuller comparisons.

\section{Conclusion}
\label{sec:conclusion}
Large Decision Models frame efficient decision making as transfer across schemas supplied at run time.
\dds{} broadens schema supervision, while \sprd{} trains the readout toward reasoning-derived decisions; DecisionVista tests their combined effect after method development.
The pre-registered transfer gains are concentrated in Qwen; a post-hoc comparison with validation-selected baseline yields smaller gains, with a remaining advantage on human-labelled DecisionVista-Human.
Synthetic-schema labels measure model agreement rather than human judgement (Appendix~\ref{app:limitations}).
Confidence routing provides a practical deployment pattern when a rollout--readout gap remains: accept confident readout answers and invoke reasoning rollout for uncertain cases.

\section*{Impact Statement}
Large Decision Models are used for moderation, triage, routing and eligibility-style decisions, where errors affect people. Our work makes such models more accurate on decisions they were not trained for and studies their calibration, which can support abstention and human review; it does not make them suitable for consequential decisions without oversight. Several of our evaluation sets are written and labelled by language models, whose shared conventions can differ from human judgement; we therefore report a human-labelled test set, release all model-labelled items with their blind labels, and specify a human audit protocol. The confidence router lowers the compute of reasoning-based decisions. All data are public or synthetic and contain no personal information beyond what the public datasets already include.

\bibliography{refs}

@article{yu2024distilling,
  title={Distilling System 2 into System 1},
  author={Yu, Ping and Xu, Jing and Weston, Jason and Kulikov, Ilia},
  journal={arXiv preprint arXiv:2407.06023},
  year={2024}
}

@inproceedings{zelikman2022star,
  title={{STaR}: Bootstrapping Reasoning With Reasoning},
  author={Zelikman, Eric and Wu, Yuhuai and Mu, Jesse and Goodman, Noah},
  booktitle={Advances in Neural Information Processing Systems},
  year={2022}
}

@inproceedings{wang2023selfconsistency,
  title={Self-Consistency Improves Chain of Thought Reasoning in Language Models},
  author={Wang, Xuezhi and Wei, Jason and Schuurmans, Dale and Le, Quoc and Chi, Ed and Narang, Sharan and Chowdhery, Aakanksha and Zhou, Denny},
  booktitle={International Conference on Learning Representations},
  year={2023}
}

@inproceedings{wang2023selfinstruct,
  title={Self-Instruct: Aligning Language Models with Self-Generated Instructions},
  author={Wang, Yizhong and Kordi, Yeganeh and Mishra, Swaroop and Liu, Alisa and Smith, Noah A. and Khashabi, Daniel and Hajishirzi, Hannaneh},
  booktitle={Proceedings of the 61st Annual Meeting of the Association for Computational Linguistics},
  year={2023}
}

@inproceedings{honovich2023unnatural,
  title={Unnatural Instructions: Tuning Language Models with (Almost) No Human Labor},
  author={Honovich, Or and Scialom, Thomas and Levy, Omer and Schick, Timo},
  booktitle={Proceedings of the 61st Annual Meeting of the Association for Computational Linguistics},
  year={2023}
}

@inproceedings{wei2022cot,
  title={Chain-of-Thought Prompting Elicits Reasoning in Large Language Models},
  author={Wei, Jason and Wang, Xuezhi and Schuurmans, Dale and Bosma, Maarten and Ichter, Brian and Xia, Fei and Chi, Ed and Le, Quoc and Zhou, Denny},
  booktitle={Advances in Neural Information Processing Systems},
  year={2022}
}

@inproceedings{wei2022flan,
  title={Finetuned Language Models are Zero-Shot Learners},
  author={Wei, Jason and Bosma, Maarten and Zhao, Vincent Y. and Guu, Kelvin and Yu, Adams Wei and Lester, Brian and Du, Nan and Dai, Andrew M. and Le, Quoc V.},
  booktitle={International Conference on Learning Representations},
  year={2022}
}

@inproceedings{sanh2022t0,
  title={Multitask Prompted Training Enables Zero-Shot Task Generalization},
  author={Sanh, Victor and Webson, Albert and Raffel, Colin and Bach, Stephen H. and others},
  booktitle={International Conference on Learning Representations},
  year={2022}
}

@inproceedings{wang2022supernatural,
  title={Super-{N}atural{I}nstructions: Generalization via Declarative Instructions on 1600+ {NLP} Tasks},
  author={Wang, Yizhong and Mishra, Swaroop and Alipoormolabashi, Pegah and Kordi, Yeganeh and others},
  booktitle={Proceedings of the 2022 Conference on Empirical Methods in Natural Language Processing},
  year={2022}
}

@inproceedings{wortsman2022wiseft,
  title={Robust Fine-Tuning of Zero-Shot Models},
  author={Wortsman, Mitchell and Ilharco, Gabriel and Kim, Jong Wook and Li, Mike and Kornblith, Simon and Roelofs, Rebecca and Lopes, Raphael Gontijo and Hajishirzi, Hannaneh and Farhadi, Ali and Namkoong, Hongseok and Schmidt, Ludwig},
  booktitle={Proceedings of the IEEE/CVF Conference on Computer Vision and Pattern Recognition},
  year={2022}
}

@inproceedings{hu2022lora,
  title={{LoRA}: Low-Rank Adaptation of Large Language Models},
  author={Hu, Edward J. and Shen, Yelong and Wallis, Phillip and Allen-Zhu, Zeyuan and Li, Yuanzhi and Wang, Shean and Wang, Lu and Chen, Weizhu},
  booktitle={International Conference on Learning Representations},
  year={2022}
}

@inproceedings{huang2023selfimprove,
  title={Large Language Models Can Self-Improve},
  author={Huang, Jiaxin and Gu, Shixiang and Hou, Le and Wu, Yuexin and Wang, Xuezhi and Yu, Hongkun and Han, Jiawei},
  booktitle={Proceedings of the 2023 Conference on Empirical Methods in Natural Language Processing},
  year={2023}
}

@inproceedings{devlin2019bert,
  title={{BERT}: Pre-training of Deep Bidirectional Transformers for Language Understanding},
  author={Devlin, Jacob and Chang, Ming-Wei and Lee, Kenton and Toutanova, Kristina},
  booktitle={Proceedings of the 2019 Conference of the North American Chapter of the Association for Computational Linguistics: Human Language Technologies, Volume 1 (Long and Short Papers)},
  pages={4171--4186},
  year={2019},
  url={https://aclanthology.org/N19-1423/}
}

@inproceedings{halder2020tars,
  title={Task-Aware Representation of Sentences for Generic Text Classification},
  author={Halder, Kishaloy and Akbik, Alan and Krapac, Josip and Vollgraf, Roland},
  booktitle={Proceedings of the 28th International Conference on Computational Linguistics},
  pages={3202--3213},
  year={2020},
  url={https://aclanthology.org/2020.coling-main.285/}
}

@inproceedings{schick2021pet,
  title={Exploiting Cloze-Questions for Few-Shot Text Classification and Natural Language Inference},
  author={Schick, Timo and Sch{\"u}tze, Hinrich},
  booktitle={Proceedings of the 16th Conference of the European Chapter of the Association for Computational Linguistics},
  year={2021}
}

@inproceedings{hsieh2023stepbystep,
  title={Distilling Step-by-Step! Outperforming Larger Language Models with Less Training Data and Smaller Model Sizes},
  author={Hsieh, Cheng-Yu and Li, Chun-Liang and Yeh, Chih-Kuan and Nakhost, Hootan and Fujii, Yasuhisa and Ratner, Alex and Krishna, Ranjay and Lee, Chen-Yu and Pfister, Tomas},
  booktitle={Findings of the Association for Computational Linguistics: ACL 2023},
  year={2023}
}

@inproceedings{ho2023teachers,
  title={Large Language Models Are Reasoning Teachers},
  author={Ho, Namgyu and Schmid, Laura and Yun, Se-Young},
  booktitle={Proceedings of the 61st Annual Meeting of the Association for Computational Linguistics},
  year={2023}
}

@inproceedings{magister2023teaching,
  title={Teaching Small Language Models to Reason},
  author={Magister, Lucie Charlotte and Mallinson, Jonathan and Adamek, Jakub and Malmi, Eric and Severyn, Aliaksei},
  booktitle={Proceedings of the 61st Annual Meeting of the Association for Computational Linguistics (Volume 2: Short Papers)},
  year={2023}
}

@article{deng2023implicit,
  title={Implicit Chain of Thought Reasoning via Knowledge Distillation},
  author={Deng, Yuntian and Prasad, Kiran and Fernandez, Roland and Smolensky, Paul and Chaudhary, Vishrav and Shieber, Stuart},
  journal={arXiv preprint arXiv:2311.01460},
  year={2023}
}

@article{deng2024explicit,
  title={From Explicit {CoT} to Implicit {CoT}: Learning to Internalize {CoT} Step by Step},
  author={Deng, Yuntian and Choi, Yejin and Shieber, Stuart},
  journal={arXiv preprint arXiv:2405.14838},
  year={2024}
}

@inproceedings{xie2020noisy,
  title={Self-Training With Noisy Student Improves {ImageNet} Classification},
  author={Xie, Qizhe and Luong, Minh-Thang and Hovy, Eduard and Le, Quoc V.},
  booktitle={Proceedings of the IEEE/CVF Conference on Computer Vision and Pattern Recognition},
  year={2020}
}

@inproceedings{lee2013pseudo,
  title={Pseudo-Label: The Simple and Efficient Semi-Supervised Learning Method for Deep Neural Networks},
  author={Lee, Dong-Hyun},
  booktitle={ICML Workshop on Challenges in Representation Learning},
  year={2013}
}

@inproceedings{zhao2021calibrate,
  title={Calibrate Before Use: Improving Few-Shot Performance of Language Models},
  author={Zhao, Zihao and Wallace, Eric and Feng, Shi and Klein, Dan and Singh, Sameer},
  booktitle={Proceedings of the 38th International Conference on Machine Learning},
  year={2021}
}

@inproceedings{kojima2022zeroshot,
  title={Large Language Models are Zero-Shot Reasoners},
  author={Kojima, Takeshi and Gu, Shixiang Shane and Reid, Machel and Matsuo, Yutaka and Iwasawa, Yusuke},
  booktitle={Advances in Neural Information Processing Systems},
  year={2022}
}

@misc{nimble,
  title={Nimble: open single-pass decision models},
  author={{Bespoke Labs}},
  howpublished={Model release and public benchmark},
  year={2026}
}

@misc{openjev,
  title={{OpenJev}: an open-weights System-One decision model},
  author={{OpenJev authors}},
  howpublished={Model release},
  year={2026}
}

@misc{jev,
  title={Jev: typed System-One decisions as a service},
  author={{TypeSafe}},
  howpublished={Hosted API, version 1.13.0},
  year={2026}
}

@inproceedings{longpre2023flan,
  title={The Flan Collection: Designing Data and Methods for Effective Instruction Tuning},
  author={Longpre, Shayne and Hou, Le and Vu, Tu and Webson, Albert and Chung, Hyung Won and Tay, Yi and Zhou, Denny and Le, Quoc V. and Zoph, Barret and Wei, Jason and Roberts, Adam},
  booktitle={Proceedings of the 40th International Conference on Machine Learning},
  year={2023}
}

@inproceedings{zhou2023lima,
  title={{LIMA}: Less Is More for Alignment},
  author={Zhou, Chunting and Liu, Pengfei and Xu, Puxin and Iyer, Srini and Sun, Jiao and Mao, Yuning and Ma, Xuezhe and Efrat, Avia and Yu, Ping and Yu, Lili and Zhang, Susan and Ghosh, Gargi and Lewis, Mike and Zettlemoyer, Luke and Levy, Omer},
  booktitle={Advances in Neural Information Processing Systems},
  year={2023}
}

@misc{laya,
  title = {Laya: A Fast, Non-Autoregressive Decision Engine with Calibrated Probabilities},
  author = {{Convai Innovations}},
  year = {2026},
  howpublished = {\url{https://huggingface.co/convaiinnovations/laya}, Apache-2.0}
}

@misc{kev,
  title = {Kev: Calibrated Typed Decision Models},
  author = {Jared Palmer},
  year = {2026},
  howpublished = {\url{https://github.com/jaredpalmer/kev}}
}

@inproceedings{guo2017calibration,
  title={On Calibration of Modern Neural Networks},
  author={Guo, Chuan and Pleiss, Geoff and Sun, Yu and Weinberger, Kilian Q.},
  booktitle={Proceedings of the 34th International Conference on Machine Learning},
  year={2017}
}

@article{chen2023frugalgpt,
  title={{FrugalGPT}: How to Use Large Language Models While Reducing Cost and Improving Performance},
  author={Chen, Lingjiao and Zaharia, Matei and Zou, James},
  journal={arXiv preprint arXiv:2305.05176},
  year={2023}
}

@article{ong2024routellm,
  title={{RouteLLM}: Learning to Route {LLMs} with Preference Data},
  author={Ong, Isaac and Almahairi, Amjad and Wu, Vincent and Chiang, Wei-Lin and Wu, Tianhao and Gonzalez, Joseph E. and Kadous, M. Waleed and Stoica, Ion},
  journal={arXiv preprint arXiv:2406.18665},
  year={2024}
}

@inproceedings{agarwal2024onpolicy,
  title={On-Policy Distillation of Language Models: Learning from Self-Generated Mistakes},
  author={Agarwal, Rishabh and Vieillard, Nino and Zhou, Yongchao and Stanczyk, Piotr and Ramos, Sabela and Geist, Matthieu and Bachem, Olivier},
  booktitle={International Conference on Learning Representations},
  year={2024},
  url={https://arxiv.org/abs/2306.13649}
}

@article{zhao2026opsd,
  title={Self-Distilled Reasoner: On-Policy Self-Distillation for Large Language Models},
  author={Zhao, Siyan and Xie, Zhihui and Liu, Mengchen and Huang, Jing and Pang, Guan and Chen, Feiyu and Grover, Aditya},
  journal={arXiv preprint arXiv:2601.18734},
  year={2026},
  url={https://arxiv.org/abs/2601.18734}
}

@inproceedings{geifman2017selective,
  title={Selective Classification for Deep Neural Networks},
  author={Geifman, Yonatan and El-Yaniv, Ran},
  booktitle={Advances in Neural Information Processing Systems},
  year={2017}
}
\bibliographystyle{icml2026}

\newpage
\appendix
\onecolumn
\section{Limitations}
\label{app:limitations}
\textbf{Model-labelled evaluation.} Our new-schema sets are labelled by agreement between models, not by humans. Part of \sprd{}'s gain on the earlier sets sits on items a third model family disputes (\S\ref{sec:analysis}); DV-Pilot and DecisionVista-Synth therefore require multi-family agreement, but agreement is not correctness. The human-labelled DecisionVista-Human confirms the direction of the effect (+1.5), but not that DecisionVista-Synth's labels are correct. We release a blinded human audit protocol whose pre-registered sample (items, sampling seed, sensitivity analysis) was fixed before the DecisionVista-Synth read; the audit is not part of this paper.
\textbf{Comparator and post-hoc analyses.} The pre-registered comparator, baseline trained for the full recipe's number of steps, is overtrained; against a baseline model selected on the validation suite the DecisionVista-Synth gain is not significant (+0.8 [$-$0.1, 1.7]). The validation-selected, one-pass, label-source, permutation and in-distribution comparisons were run after the sealed reads and are descriptive.
\textbf{Novelty is screened, not guaranteed.} DecisionVista-Synth families were screened against all training schemas and earlier sets by model audit and embedding similarity; they may still resemble pretraining data.
\textbf{Author--family confounding.} Each DecisionVista-Synth family has one author, so author-level results mix author style with family difficulty.
\textbf{Scope.} We study English decisions with up to eight options, training contexts under 2k tokens, LoRA adapters, and three base models up to 31B that differ in family, size and post-training, so differences between them are not a controlled scaling study; we do not claim the results for full fine-tuning, long contexts, multilingual inputs or larger models. Gemma gains least; relabelling its \sprd{}-C pool with its own reasoning gives equivalent results (Appendix~\ref{app:gself}), so teacher mismatch does not explain this, and the strength of its untuned readout is the likelier cause. The self-posed items are written by Qwen3.8-27B for every base.
\textbf{Calibration of Gemma.} All Gemma variants, including step-matched baseline, remain overconfident (transfer-suite ECE 21--22); its fitted temperatures are 10--12 in every variant, and our recipe does not fix this; Gemma \ldm{}s need post-hoc temperature scaling.
\textbf{Router cost.} Router savings are in generated tokens; routed items also need the reasoning model's prefill and a second model in serving, and we did not measure end-to-end routing latency or throughput.
\textbf{Compute.} The \sprd{} teacher pass costs about 8 H100-hours per 16k items; seeds beyond three were not affordable for the 27B and 31B variants.

\section{Supervised recipes behind the diagnosis}
\label{app:recipes}
Table~\ref{tab:recipes} lists every supervised recipe we ran before \sprd{}, with one seed each unless marked. All use the readout objective (cross-entropy over option letters with per-epoch option shuffling) and LoRA rank 16 unless marked.

\begin{table*}[tp]
\centering\small
\caption{Supervised recipes run before \sprd{}. None improves new-schema accuracy (DV-Dev-A, DV-Dev-B) of the 27B base beyond seed noise; post-hoc update scaling recovers some. The eight-subset diagnostic averages the choice and yes/no subsets of Nimble, excluding Civil Comments and MultiNLI.}
\label{tab:recipes}
\setlength{\tabcolsep}{3pt}
\fitboxw{%
\begin{tabular}{lllrrrr}
\toprule
Base & Recipe & Change & 8 subsets & Val. & DV-Dev-A & DV-Dev-B \\
\midrule
27B & untuned & -- & 82.5 & 79.3 & 89.1 & 87.5 \\
27B & Baseline & 20k public-dataset records + Nimble data & 82.6 & 81.8 & 88.1 & 87.5 \\
27B & Baseline, full fine-tune & all weights, ZeRO-3 & 80.6 & 78.1 & 87.1 & -- \\
27B & Baseline + source splits & + 21k benchmark-source train records & 89.1 & 82.6 & 87.8 & 88.0 \\
27B & Baseline+adversarial pairs & + 6.5k LLM minimal pairs, flip margin & 83.7 & 82.0 & 89.1 & 88.3 \\
27B & Baseline, update $\times0.5$ & post-hoc scaling & 82.8 & 82.4 & 89.9 & 87.5 \\
27B & Baseline + source splits, update $\times0.5$ & post-hoc scaling & 88.0 & 82.0 & 89.7 & 89.2 \\
\midrule
9B & untuned & -- & 77.6 & 71.7 & 79.9 & 74.8 \\
9B & Nimble data only & Nimble's recipe & 80.8 & 72.9 & -- & -- \\
9B & Baseline (3 seeds) & & 80.4--81.3 & 75.6--77.3 & 84.3--85.5 & 79.1 \\
9B & + permutation consistency & JS between two option orders & 80.9 & 77.9 & 86.5 & -- \\
9B & + ordinal loss & ranked-probability score on scores & 80.3 & 75.8 & 83.5 & -- \\
9B & + self-supervised decisions & 12k denoising decisions & 81.3 & 76.5 & 84.5 & -- \\
9B & + distillation from OpenJev & KL to OpenJev readout & 81.8 & 76.8 & 86.9 & -- \\
9B & + synthetic task-matched & 4k LLM items for suite tasks & 81.1 & 77.5 & 87.6 & -- \\
9B & + executable curriculum & 12k program-labelled items & 79.0 & 76.3 & 85.2 & 79.9 \\
9B & + retention KL + adversarial pairs & & 80.9 & 77.1 & 86.6 & 81.9 \\
9B & rank 64, 3 epochs & capacity & 80.0 & 75.4 & 84.5 & -- \\
9B & Baseline + source splits & & 88.2 & 76.0 & 84.7 & -- \\
\midrule
\multicolumn{3}{l}{OpenJev-27B (released)} & 84.1 & 84.1 & 88.6 & 86.1 \\
\multicolumn{3}{l}{Nimble-9B (released)} & 81.6 & 73.0 & 82.7 & 79.7 \\
\bottomrule
\end{tabular}}
\end{table*}

Halving the learned baseline update after training~\citep{wortsman2022wiseft} recovers 1.8 points on DV-Dev-A (88.1 to 89.9), a diagnostic motivation for reporting update-scaled baseline comparisons.

\section{Statistics}
\label{app:stats}
\paragraph{Recipe comparisons (Tables~\ref{tab:contrast}, \ref{tab:fresh5}).} For base $b$, recipe $r$ and training seed $s$, let $A_{b,r,s}$ be an accuracy statistic. The effect of $r$ at base $b$ is $\Delta_b = \frac1{|S_b|}\sum_{s\in S_b}(A_{b,r,s}-A_{b,\text{baseline},s})$: each seed of the recipe is paired with the baseline variant of the \emph{same} seed. The pooled effect is the unweighted mean over bases, $\Delta=\frac13\sum_b\Delta_b$. Seeds: $S_{9B}=\{17,29,41\}$ everywhere; $S_{27B}=S_{31B}=\{17,29\}$ for the test sets (as pre-registered) and $\{17,29,41\}$ for validation sets.
\paragraph{Statistics per set.} Validation: $A$ is the unweighted mean of the pooled item accuracies on DV-Dev-A, DV-Dev-B and the transfer suite. DV-Pilot and DecisionVista-Synth: family-macro accuracy (families weighted equally). DecisionVista-Human: source-macro accuracy (eight sources weighted equally).
\paragraph{Intervals.} We resample the unit that defines novelty, with the same resample indices for every variant, seed and base, so that differences are paired: task families for the synthetic DecisionVista sets (contrastive groups stay intact), families and held-out source subjects for the exploratory statistic (each of the three sets resampled separately), and items within each source for DecisionVista-Human, whose eight sources are fixed. 10,000 resamples; percentile intervals; two-sided bootstrap $p$-values. For DecisionVista-Synth we also report the pre-registered sign-flip test over family-level differences (100,000 flips). The intervals condition on the trained models; seed variation is shown separately by the per-seed effects (all 27 exploratory per-seed differences are positive; DecisionVista-Synth per-seed effects at 9B are +2.7, +4.7 and $-$0.6, at 27B +2.1 and +2.6, and at 31B +0.6 and +0.8).
\paragraph{Multiplicity.} Only the two primary contrasts are confirmatory. On DecisionVista-Synth the per-scale tests are Holm-corrected and gatekept by the primary, followed by the ordered component tests; all other cells are descriptive.
\paragraph{Earlier comparisons.} Analyses of the \sprd{}-only experiments (DV-Dev-A--DV-Pilot, Table~\ref{tab:samesubset} and \S\ref{sec:analysis}) use a paired bootstrap over generation batches or task families with 5,000--20,000 resamples, as reported with each result.

\section{Protocol chronology}
\label{app:chronology}
Table~\ref{tab:chrono} lists every decision that touched a test set. The time-stamped planning log in which each entry was written before the corresponding read, including the DV-Pilot file hash, is part of the released artifacts. DV-Dev-A, DV-Dev-B and DV-Stress are validation evidence; DV-Pilot is confirmatory for \sprd{}-C and DecisionVista-Synth for the combined recipe.
\begin{table}[h]
\centering\scriptsize
\caption{Order of decisions relative to test-set reads.}
\label{tab:chrono}
\begin{tabular}{p{0.9cm}p{5.8cm}}
\toprule
Step & Event \\
\midrule
1 & DV-Dev-A and DV-Dev-B built and used for earlier recipes (Appendix~\ref{app:recipes}); DV-Stress built and sealed. \\
2 & Pool generated; DV-Stress family descriptions used only in the exclusion list. \\
3 & $\lambda$, $\tau$ chosen on the validation suite and DV-Dev-A (9B); \sprd{} variants and DV-Stress checkpoints frozen and recorded. \\
4 & DV-Stress read once (+1.0 over baseline). \\
5 & Third-family (Magistral) labels for DV-Dev-A, DV-Dev-B and DV-Stress; gains concentrate on contested items. \\
6 & Step-matched baseline, \sprd{}-C and other variants trained; DV-Pilot families proposed and audited, protocol recorded. \\
7 & DV-Pilot items generated and sealed after three-way agreement (file hash recorded). \\
8 & Variant chosen on the validation suite, DV-Dev-A, DV-Dev-B and validated subsets (\sprd{}-C) and recorded. \\
9 & DV-Pilot read once for the \sprd{} rows. \\
10 & \dds{} mixture chosen after a 4B screen scored on the validation suite, DV-Dev-A, DV-Dev-B, the eight-subset diagnostic and the transfer suite; models trained (seeds 17/29 at every base, 41 at 9B; never read on DV-Stress, DV-Pilot or DV-Synth). \\
11 & DV-Synth designed with advisors; families built by decision operator from four authors (three not used in any training data); pre-registration written (primary contrast \dds{}+\sprd{}-C vs.\ step-matched baseline), then amended to place the human audit at the end. \\
12 & DV-Synth items generated and labelled; hashes recorded. \\
13 & Seed 41 at 27B and 31B trained (after the pre-registration, before the read); variant manifest (adapter hashes) and analysis code frozen; DV-Synth read once for all variants and references. \\
14 & After internal review, DV-Human built from datasets no variant or design decision had used, hashed and pre-registered, then read once for all variants. \\
15 & Post-hoc, descriptive: one-pass baselines at 27B and 31B; Laya references. \\
16 & Post-hoc amendment for uniform labels: DV-Dev-A, DV-Dev-B and DV-Stress relabelled blind by GLM-5.2, DeepSeek-V4 and Kimi-K3 with DV-Synth's prompt; items kept iff at least two of three agree with the author (2,311 of 2,374, 97.3\%); all variants re-scored from stored predictions. The factorial's development contrasts and score tables use these kept items; the diagnosis, recipe-study and routing analyses report the full sets. \\
\bottomrule
\end{tabular}
\end{table}

\section{Cost}
\label{app:cost}
The teacher pass over the 16k-item pool generated about 59M tokens (a median of 2,051 tokens per item for four samples) and took about 8 GPU-hours on H100s with vLLM in FP8; schema and item generation took under 2 GPU-hours.
In the earlier \sprd{} study the 14.8k pool records made training about 1.7$\times$ as long as baseline training (4.6 vs.\ 2.8 hours for 27B on one H100); in the controlled study the full recipe and step-matched baseline train the same budget (4,031 and 4,032 steps), and the single-component variants fewer.
Inference is unchanged: one prefill and no generated tokens, against a median of 335--513 generated tokens per sample for the reasoning rollout that the student imitates.

\section{Additional analysis tables}
\label{app:extra}
\begin{table}[h]
\centering\small
\caption{Rollout--readout gap of untuned Qwen3.8-27B. Reasoning = majority of four thinking-mode samples; readout = single-pass letter readout.}
\label{tab:devgap}
\begin{tabular}{lrrr}
\toprule
Validation source & Readout & Reasoning & Gap \\
\midrule
WinoGrande & 67.5 & 93.5 & +26.0 \\
MedMCQA & 65.0 & 72.5 & +7.5 \\
TweetEval irony & 69.5 & 76.0 & +6.5 \\
SMS spam & 97.5 & 98.5 & +1.0 \\
ARC-Challenge & 96.0 & 96.5 & +0.5 \\
CaseHOLD & 78.5 & 78.0 & $-0.5$ \\
CommitmentBank & 85.7 & 69.6 & $-16.1$ \\
\midrule
Val. (all, 1,256) & 79.3 & 85.1 & +5.8 \\
DV-Dev-A (613) & 89.1 & 96.9 & +7.8 \\
DV-Dev-B (592) & 87.5 & 96.8 & +9.3 \\
\bottomrule
\end{tabular}
\end{table}
\begin{table}[h]
\centering\small
\caption{Agreement (\%) of other label sources with the reasoning-teacher majority on the self-posed pool, by the teacher's vote share (15,980 of the 16,040 items have at least one parseable teacher answer).}
\label{tab:agree}
\begin{tabular}{lrrrr}
\toprule
Vote share & Items & Readout & gpt-5.5 & Gen.\ intent \\
\midrule
1.0 & 13,191 & 85.9 & 93.9 & 73.9 \\
$[0.75, 1)$ & 1,638 & 54.8 & 64.0 & 43.2 \\
$<0.75$ & 1,151 & 47.3 & 51.9 & 40.0 \\
\midrule
All & 15,980 & 79.9 & 87.8 & 68.3 \\
\bottomrule
\end{tabular}
\end{table}

\section{All scores}
\label{app:allscores}
\begin{table*}[tp]
\centering\small
\caption{Accuracy (\%) and ECE (\%, pooled over Val., DV-Dev-A, DV-Dev-B and the transfer suite; lower is better); seed means. Best per column within each block in bold; our recipe shaded. The eight-subset diagnostic averages Nimble's choice and yes/no subsets excluding Civil Comments and MultiNLI. It, Nimble-13 (the full benchmark), Val. and the transfer suite are human-labelled; synthetic DecisionVista sets are new decision schemas labelled by model agreement. No tuned row trains on any benchmark source. Val., the transfer suite, DV-Dev-A and DV-Dev-B informed design or selection decisions (validation evidence); DV-Pilot, DecisionVista-Synth (family-macro) and DecisionVista-Human (DV-Human, source-macro; human-labelled) were sealed and read once for all rows after the variants were frozen; before that final read, no \dds{} variant had been scored on DV-Pilot. $^\ddagger$Fine-tuned by its authors on the training split of their typed-decisions benchmark.}
\label{tab:main}
\setlength{\tabcolsep}{2.6pt}
\fitboxw{%
\begin{tabular}{lrrrrrrrrrrr}
\toprule
 & & \multicolumn{2}{c}{Public benchmarks} & \multicolumn{4}{c}{Validation} & \multicolumn{3}{c}{Test} & \\
\cmidrule(lr){3-4}\cmidrule(lr){5-8}\cmidrule(lr){9-11}
Model & Seeds & 8 subsets & Nimble-13 & Val. & Transfer & \shortstack{DV-\\Dev-A} & \shortstack{DV-\\Dev-B} & \shortstack{DV-\\Pilot} & \shortstack{DV-\\Synth} & \shortstack{DV-\\Human} & ECE$\downarrow$ \\
\midrule
\multicolumn{12}{l}{\emph{Released decision encoders}} \\
Laya-en (0.4B) & 1 & 61.3 & 57.2 & 44.3 & 35.0 & 54.6 & 44.1 & 54.2 & 33.3 & -- & 17.8 \\
Laya-multilingual (0.3B) & 1 & \textbf{63.3} & 57.9 & 40.0 & 30.8 & 51.4 & 38.0 & 45.0 & 30.8 & -- & 30.5 \\
Laya-typed-decisions$^\ddagger$ (0.4B) & 1 & 61.1 & \textbf{59.8} & \textbf{48.3} & \textbf{35.5} & \textbf{55.6} & \textbf{46.1} & \textbf{57.9} & \textbf{33.8} & -- & \textbf{7.3} \\
\midrule
\multicolumn{12}{l}{\emph{Qwen3.5-9B}} \\
Untuned & 1 & 77.3 & 73.7 & 71.7 & 66.2 & 80.5 & 75.6 & 90.8 & 52.5 & 69.4 & 14.0 \\
Nimble-9B & 1 & \textbf{81.7} & 74.8 & 72.7 & 67.5 & 83.5 & 82.2 & 93.2 & 63.6 & 72.5 & 12.4 \\
Baseline, step-matched & 3 & 80.0 & 74.1 & 75.9 & 69.7 & 84.1 & 79.3 & 91.8 & 64.5 & 72.8 & 20.3 \\
\rowcolor{gray!15} + \dds{} & 3 & 80.8 & 71.6 & 77.1 & 71.6 & 87.2 & 83.8 & 96.6 & 66.1 & \textbf{75.0} & \textbf{7.5} \\
\rowcolor{gray!15} + \sprd{}-C & 3 & 81.2 & 76.1 & \textbf{77.4} & 71.1 & \textbf{87.6} & 84.0 & 96.1 & 66.5 & 74.9 & 9.4 \\
\rowcolor{gray!15} + \dds{} + \sprd{}-C & 3 & 80.7 & \textbf{76.2} & 77.1 & \textbf{71.9} & 87.1 & \textbf{85.8} & \textbf{96.8} & \textbf{66.8} & 74.9 & 7.8 \\
\midrule
\multicolumn{12}{l}{\emph{Qwen3.8-27B}} \\
Untuned & 1 & 82.5 & 74.4 & 79.3 & 75.4 & 89.2 & 88.8 & 97.6 & 68.3 & 73.2 & \textbf{3.1} \\
OpenJev-27B & 1 & \textbf{83.9} & \textbf{77.7} & 83.0 & 76.5 & 90.0 & 89.1 & 98.3 & 75.3 & 76.5 & 4.5 \\
Baseline, step-matched & 3 & 82.0 & 75.2 & 81.6 & 76.2 & 89.4 & 87.1 & 97.1 & 76.1 & 75.2 & 15.0 \\
\rowcolor{gray!15} + \dds{} & 3 & 82.3 & 74.9 & 83.3 & 77.4 & 90.5 & 89.6 & 98.5 & 77.7 & 76.3 & 5.7 \\
\rowcolor{gray!15} + \sprd{}-C & 3 & 83.1 & 76.3 & 82.9 & 76.9 & \textbf{91.2} & 90.3 & \textbf{98.9} & 77.9 & 76.7 & 8.2 \\
\rowcolor{gray!15} + \dds{} + \sprd{}-C & 3 & 83.0 & 75.7 & \textbf{83.7} & \textbf{77.5} & 91.1 & \textbf{91.1} & 98.7 & \textbf{78.2} & \textbf{77.0} & 6.5 \\
\midrule
\multicolumn{12}{l}{\emph{Gemma-4-31B}} \\
Untuned & 1 & \textbf{84.3} & \textbf{77.8} & 82.3 & 77.8 & 90.8 & 90.3 & 98.9 & 77.2 & 79.2 & 18.3 \\
Baseline, step-matched & 3 & 82.1 & 75.3 & 82.4 & 77.6 & 89.5 & 87.3 & 97.9 & 79.9 & 79.5 & 18.8 \\
\rowcolor{gray!15} + \dds{} & 3 & 82.6 & 75.6 & 83.6 & 78.9 & 91.1 & 89.8 & 99.3 & \textbf{81.0} & 79.5 & 17.2 \\
\rowcolor{gray!15} + \sprd{}-C & 3 & 83.0 & 76.6 & 83.2 & 78.0 & \textbf{91.4} & 90.6 & 99.3 & 79.6 & 79.4 & 17.7 \\
\rowcolor{gray!15} + \dds{} + \sprd{}-C & 3 & 83.4 & 76.3 & \textbf{84.1} & \textbf{79.0} & 91.3 & \textbf{91.6} & \textbf{99.4} & 80.8 & \textbf{80.2} & \textbf{16.9} \\
\bottomrule
\end{tabular}}
\end{table*}

\section{Training configurations}
\label{app:spec}
Everything in Tables~\ref{tab:released}, \ref{tab:contrast}, \ref{tab:fresh5} and \ref{tab:main} comes from the variants below; the earlier recipe study (Appendix~\ref{app:recipes}) and the 4B \dds{} ablations used other settings and are reported separately.
\begin{table}[h]
\centering\scriptsize
\caption{Configuration of every variant in the controlled study.}
\label{tab:spec}
\setlength{\tabcolsep}{3pt}
\begin{tabular}{@{}p{0.3\columnwidth}p{0.64\columnwidth}@{}}
\toprule
Item & Setting \\
\midrule
Bases & Qwen3.5-9B; Qwen3.8-27B (rev.\ 1d4bf0f2); Gemma-4-31B-it \\
Adapter & LoRA rank 16, $\alpha{=}32$, dropout 0.05, on every linear layer of the language model; bf16; final step, no checkpoint selection \\
Optimiser & AdamW, lr $5{\times}10^{-5}$, linear schedule with 5\% warm-up, gradient clipping 1.0, effective batch 16 (9B: 4$\times$4 accumulation; 27B/31B: 2$\times$8), max length 2,048 \\
Objective & Eq.~\ref{eq:loss}: cross-entropy on baseline and \dds{} records; $\lambda{=}3$ KL to \sprd{}-C vote targets; option order re-shuffled per draw \\
Data & baseline = 20k public records + Nimble's 2.7k training items (22.4k); \dds{} = half of the 63.4k mixture (31.7k, fixed sample); \sprd{}-C = 10,331 items labelled by Qwen3.8-27B for every base \\
Steps & baseline, step-matched: 4,032 (2.9 passes); +\dds{}: 3,386; +\sprd{}-C: 2,050; +both: 4,031 (one pass each); one-pass baseline: 1,405 \\
Seeds & 17, 29, 41 (seed 41 at 27B/31B trained after the DV-Synth pre-registration) \\
Readout & Eq.~\ref{eq:readout} with Nimble's scorer, canonical option order at test time, argmax, max input 8,192 tokens \\
Manifest & adapter SHA-256 of all 36 variants, analysis scripts and set hashes are frozen in the released MANIFEST file \\
\bottomrule
\end{tabular}
\end{table}

\paragraph{Experiment summary.} Table~\ref{tab:ledger} maps every trained configuration that appears in a table to its data, budget, seeds and analysis status. ``Pre-reg.'' rows are covered by the DecisionVista-Synth and DecisionVista-Human pre-registrations; all other comparisons on the test sets are post hoc.
\begin{table}[h]
\centering\scriptsize
\caption{Experiment summary. Steps at effective batch 16. Seeds: 9B / 27B / 31B. Status: \emph{pre-reg.}, primary or secondary contrast fixed before the sealed reads; \emph{descr.}, descriptive; \emph{post hoc}, specified after the sealed reads.}
\label{tab:ledger}
\setlength{\tabcolsep}{3pt}
\begin{tabular}{@{}llrlll@{}}
\toprule
Configuration & Training data & Steps & Seeds & Status & Where \\
\midrule
Baseline, step-matched & baseline 22.4k & 4,032 & 3 / 3 / 3 & pre-reg.\ comparator & Tables~\ref{tab:released},~\ref{tab:contrast},~\ref{tab:main} \\
+\dds{} + \sprd{}-C & baseline + \dds{} 31.7k + \sprd{}-C 10.3k & 4,031 & 3 / 3 / 3 & pre-reg.\ primary & Tables~\ref{tab:released},~\ref{tab:contrast},~\ref{tab:main} \\
+\dds{} & baseline + \dds{} & 3,386 & 3 / 3 / 3 & pre-reg.\ secondary; descr.\ vs.\ baseline & Tables~\ref{tab:contrast},~\ref{tab:main} \\
+\sprd{}-C & baseline + \sprd{}-C & 2,050 & 3 / 3 / 3 & pre-reg.\ secondary; descr.\ vs.\ baseline & Tables~\ref{tab:contrast},~\ref{tab:main} \\
Baseline, one pass & baseline & 1,405 & 2 / 2 / 2 & descr. & Table~\ref{tab:main} \\
Validation-selected baseline and recipe & as above, update scale 1 or 0.65 & -- & 3 / 2 / 2 & post hoc & Tables~\ref{tab:contrast},~\ref{tab:devsel} \\
+\dds{} or +\sprd{}-C, 4,031 steps & as the individual components & 4,031 & 3 / 2 / -- & post hoc & Table~\ref{tab:stepcomp} \\
Pool with generator labels & baseline + 10,331 random pool items & 2,050 & 3 / -- / -- & post hoc & \S\ref{sec:analysis} \\
Label-source controls & baseline + 14.8k pool, per label source & 1.66 passes & -- / 2 / -- & post hoc (earlier study) & Table~\ref{tab:labelctl} \\
+ benchmark source splits & recipe + public-suite training splits & -- & 1 / 2 / -- & descr. & Table~\ref{tab:indist} \\
Confidence router & no training; validation threshold & -- & as recipe & post hoc on test sets & Fig.~\ref{fig:router}, Table~\ref{tab:router} \\
\bottomrule
\end{tabular}
\end{table}

\begin{table*}[tp]
\centering\scriptsize
\caption{Nimble's 13-subset public benchmark (accuracy, \%; three-seed means for tuned rows, seed standard deviation shown when $\geq$3). No tuned row trains on any of these sources. HelpSteer2 and SummEval are 5-level ratings; SummEval-consistency (144 items) is unstable across seeds of the same recipe (13--84), which dominates the seed variance of the 13-subset average. Avg-13 is the full-benchmark score. Avg-11 is an exploratory sensitivity average excluding Civil Comments (89\% majority class) and MultiNLI, retained to show dependence on these two subsets. Jev 1.13.0's published 13-subset average is 76.0.}
\label{tab:public13}
\setlength{\tabcolsep}{2.2pt}
\fitboxw{%
\begin{tabular}{lrrrrrrrrrrrrrrr}
\toprule
Model & VitC & MAS-en & MAS-de & PubMed & BoolQ & SQuAD2 & PAWS & Aegis2 & MNLI & CivilC & HelpS2 & SE-rel & SE-con & Avg-13 & Avg-11 \\
\midrule
\multicolumn{16}{l}{\emph{Qwen3.5-9B}} \\
Untuned & 75.3 & 84.0 & 78.3 & 65.6 & 80.3 & \textbf{84.0} & 75.2 & 76.8 & 86.0 & 80.0 & \textbf{38.5} & 51.7 & \textbf{82.6} & 73.7 & 72.0 \\
Nimble-9B & \textbf{76.6} & 86.9 & 83.7 & 74.8 & 86.0 & 80.9 & \textbf{83.2} & 81.2 & 86.0 & 70.3 & \textbf{38.5} & 49.6 & 75.0 & 74.8 & \textbf{74.2} \\
Baseline, step-m. & 67.9 & 86.6 & 84.3 & 74.4 & \textbf{86.9} & 82.7 & 75.7 & 81.2 & \textbf{90.3} & 86.9 & 35.5 & \textbf{52.2} & 58.1\textsubscript{$\pm$30.6} & 74.1 & 71.4 \\
\rowcolor{gray!15} +\dds{} & 72.5 & \textbf{87.8} & \textbf{84.4} & \textbf{77.2} & 86.8 & 78.6 & 76.8 & 82.1 & 88.2 & \textbf{89.9} & 36.4 & 36.4\textsubscript{$\pm$8.8} & 34.3\textsubscript{$\pm$26.4} & 71.6 & 68.5 \\
\rowcolor{gray!15} +\sprd{}-C & 74.2 & 86.8 & 83.0 & 75.6 & 86.6 & 79.7 & 81.5 & \textbf{82.3} & 90.2 & 88.9 & 36.9 & 47.9\textsubscript{$\pm$3.3} & 75.9\textsubscript{$\pm$12.2} & 76.1 & 73.7 \\
\rowcolor{gray!15} +both & 70.6 & 86.9 & 82.6 & 74.8 & \textbf{86.9} & 80.7 & 81.3 & 81.6 & 88.2 & 89.6 & \textbf{38.6} & 46.9\textsubscript{$\pm$3.7} & 81.9\textsubscript{$\pm$3.0} & \textbf{76.2} & 73.9 \\
\midrule
\multicolumn{16}{l}{\emph{Qwen3.8-27B}} \\
Untuned & 75.0 & 85.7 & 85.4 & 71.2 & 86.3 & \textbf{87.6} & \textbf{85.2} & 83.2 & 84.6 & 76.3 & 38.5 & 21.2 & \textbf{86.8} & 74.4 & 73.3 \\
OpenJev-27B & \textbf{80.5} & 87.7 & 86.9 & \textbf{81.2} & 88.3 & 83.6 & 82.4 & 80.8 & 89.0 & 86.7 & \textbf{43.0} & \textbf{35.0} & 85.4 & \textbf{77.7} & \textbf{75.9} \\
Baseline, step-m. & 72.4 & 86.7 & 87.5 & 77.7 & 88.3 & 81.3 & 79.2 & 82.5 & \textbf{92.1} & 87.9 & 38.4 & 22.9\textsubscript{$\pm$4.6} & 80.3 & 75.2 & 72.5 \\
\rowcolor{gray!15} +\dds{} & 76.6 & 88.0 & \textbf{87.9} & 77.5 & 88.9 & 79.2 & 77.2 & \textbf{83.3} & 91.3 & \textbf{90.2} & 36.4 & 13.5\textsubscript{$\pm$8.9} & 84.0 & 74.9 & 72.0 \\
\rowcolor{gray!15} +\sprd{}-C & 77.8 & 87.4 & 87.6 & 78.5 & \textbf{89.6} & 81.8 & 78.5 & \textbf{83.3} & 91.4 & 89.4 & 42.4 & 21.7 & 81.7 & 76.3 & 73.7 \\
\rowcolor{gray!15} +both & 77.1 & \textbf{88.4} & 87.2 & 77.7 & 88.2 & 80.7 & 81.2 & \textbf{83.3} & \textbf{92.1} & 89.9 & 41.4\textsubscript{$\pm$3.0} & 14.6\textsubscript{$\pm$9.0} & 82.2 & 75.7 & 72.9 \\
\midrule
\multicolumn{16}{l}{\emph{Gemma-4-31B}} \\
Untuned & \textbf{77.1} & 87.7 & 89.1 & 79.2 & 88.3 & \textbf{87.0} & \textbf{85.6} & 80.0 & 87.3 & 80.3 & 42.2 & \textbf{42.5} & \textbf{85.4} & \textbf{77.8} & \textbf{76.7} \\
Baseline, step-m. & 69.6\textsubscript{$\pm$4.3} & \textbf{89.1} & \textbf{90.1} & \textbf{80.0} & 88.0 & 81.2\textsubscript{$\pm$4.0} & 74.4 & 84.3 & 90.1 & \textbf{89.0} & 39.5 & 21.7\textsubscript{$\pm$4.0} & 82.2 & 75.3 & 72.7 \\
\rowcolor{gray!15} +\dds{} & 77.0 & 88.3 & 88.8 & 78.0 & 88.3 & 80.5\textsubscript{$\pm$3.4} & 76.8 & 83.5 & 88.7 & 88.0 & 38.7\textsubscript{$\pm$4.7} & 26.4\textsubscript{$\pm$9.4} & 79.9\textsubscript{$\pm$3.6} & 75.6 & 73.3 \\
\rowcolor{gray!15} +\sprd{}-C & 75.3 & 87.6 & 88.6 & 78.9 & 87.9 & 82.4\textsubscript{$\pm$4.2} & 79.5 & 84.1 & \textbf{90.3} & 87.7 & \textbf{42.6} & 29.2\textsubscript{$\pm$6.8} & 81.9 & 76.6 & 74.4 \\
\rowcolor{gray!15} +both & 76.2 & 88.1 & 87.9 & 77.9 & \textbf{89.2} & 82.1 & 81.1 & \textbf{84.5} & 88.6 & 88.0 & 38.2\textsubscript{$\pm$4.9} & 29.6 & 80.6\textsubscript{$\pm$5.5} & 76.3 & 74.1 \\
\bottomrule
\end{tabular}}
\end{table*}

\section{Benchmark-source training comparisons}
\label{app:sourcecomparisons}
Adding the public benchmark's training splits makes the Nimble benchmark in-distribution for our source-augmented runs.
Table~\ref{tab:indist} reports this separate comparison regime, including source exposure and trainer differences.
With the controlled study trainer, \dds{} plus source splits reaches 80.8 over the full 13-subset benchmark at 9B and 81.7 at 27B, compared with Nimble-9B's 74.8 and OpenJev's 77.7.
Adding \sprd{}-C does not improve this benchmark further.
\begin{table*}[!t]
\centering\footnotesize
\caption{Models that train on some benchmark sources, compared like with like. Our rows add the training splits of the public suite's sources (so the Nimble benchmark is in-distribution for them) with the same trainer as the controlled study; Kev trains on BoolQ and MultiNLI (two public-suite subsets), Laya-typed-decisions on the typed-decisions training split, and OpenJev's training data are undisclosed. Our 9B baseline + source splits row uses the earlier recipe-study trainer. The 11-subset column is an exploratory sensitivity average excluding Civil Comments and MultiNLI (Table~\ref{tab:public13}). Best per block in bold.}
\label{tab:indist}
\setlength{\tabcolsep}{3.6pt}
\fitboxw{%
\begin{tabular}{lrrrrrr}
\toprule
 & \multicolumn{4}{c}{Public benchmarks (test)} & \multicolumn{2}{c}{Ours: development / sealed test} \\
\cmidrule(lr){2-5}\cmidrule(lr){6-7}
Model & Nimble-13 & Nimble-11 & Kev-Transfer & Typed-Decisions & Transfer suite & DV-Human \\
 & {\color{black!55}\scriptsize Nimble; 13 tasks} & {\color{black!55}\scriptsize w/o CC, MNLI} & {\color{black!55}\scriptsize Kev; new sources} & {\color{black!55}\scriptsize Laya/Jev; workflows} & {\color{black!55}\scriptsize dev.; 8 MC sources} & {\color{black!55}\scriptsize sealed; 8 sources} \\
\midrule
\multicolumn{7}{l}{\emph{Small encoder}} \\
Laya-TD, 0.4B (TD train) & 59.8 & 54.3 & 67.1 & 76.2 & 35.5 & 51.6 \\
\multicolumn{7}{l}{\emph{Qwen3.5-9B base}} \\
Kev-9B (BoolQ, MNLI) & 75.3 & 73.1 & 85.2 & 71.4 & 65.9 & 72.6 \\
Ours: baseline + splits & \textbf{82.6} & \textbf{80.9} & -- & 67.0 & 70.2 & 72.2 \\
\rowcolor{gray!15} Ours: \dds{} + splits & 80.8 & 79.6 & \textbf{85.4} & 71.6 & \textbf{71.9} & 74.4 \\
\rowcolor{gray!15} Ours: \dds{} + splits + \sprd{}-C & 80.0 & 78.5 & \textbf{85.4} & \textbf{72.7} & 71.5 & \textbf{74.7} \\
\multicolumn{7}{l}{\emph{Qwen3.8-27B base}} \\
Kev-27B (BoolQ, MNLI) & 77.8 & 75.8 & 89.5 & \textbf{75.9} & 73.5 & 75.5 \\
OpenJev-27B (undisclosed) & 77.7 & 75.9 & 88.3 & 73.0 & 76.5 & 76.5 \\
\rowcolor{gray!15} Ours: \dds{} + splits (2 seeds) & \textbf{81.7} & \textbf{80.3} & \textbf{90.7} & 74.8 & 77.6 & 76.2 \\
\rowcolor{gray!15} Ours: \dds{} + splits + \sprd{}-C & 81.6 & 80.0 & 90.5 & 75.1 & \textbf{77.8} & \textbf{76.9} \\
\bottomrule
\end{tabular}}
\end{table*}

\section{The \dds{} corpus}
\label{app:dds}
\begin{table*}[tp]
\centering\small
\caption{\dds{} components (``$\to$'': before and after the evaluation-family audit). All records are 13-gram de-contaminated against evaluation items available when the mixture was built; the later DecisionVista-Synth and DecisionVista-Human sets were checked against the frozen mixture.}
\label{tab:dds}
\setlength{\tabcolsep}{3pt}
\begin{tabular}{@{}lrp{0.6\textwidth}@{}}
\toprule
Component & Records & Content \\
\midrule
additional public data & 35.6k & 54 public sources (medicine, science, math/logic incl.\ AQuA and 26 BIG-bench tasks, reading, commonsense, preference: HH-RLHF, SHP, PKU-SafeRLHF, safety, code, fact-checking, finance) \\
Super-NaturalInstructions & 15.4k & 258 categorical tasks; tasks derived from evaluation sources excluded by name and by content (a BoolQ-derived task was removed after inspection) \\
grid-based synthesis & 13.0k $\to$ 8.4k & gpt-5.6 (sol/terra/luna) invents tasks over 44 fields $\times$ 17 decision types; kept only if gpt-6-astra agrees blind; audited against evaluation families \\
grounded synthesis & 10.7k $\to$ 3.9k & new decision questions about real public documents, audited likewise \\
\midrule
\dds{} mixture & 63.4k & additional public data + instruction tasks + audited synthesis; no benchmark-source data; half (31.7k) sampled per run \\
\bottomrule
\end{tabular}
\end{table*}
\begin{table}[t]
\centering\scriptsize
\caption{\dds{} ablations on Qwen3.5-4B (``+'': added to the 44k baseline data; seed means where $n{>}1$). Mean (3 sets) averages DV-Dev-A, DV-Dev-B and the transfer suite. The eight-subset diagnostic uses Nimble's choice and yes/no subsets excluding Civil Comments and MultiNLI. $^\dagger$This 4B recipe includes benchmark-source training splits, so the eight-subset diagnostic is in-distribution here.}
\label{tab:ddsabl}
\setlength{\tabcolsep}{2.5pt}
\fitbox{%
\begin{tabular}{lrrrrrrrr}
\toprule
Data & $n$ & Val. & DV-Dev-A & DV-Dev-B & Transfer & 8 subsets$^\dagger$ & ECE & Mean (3 sets) \\
\midrule
Baseline 44k, 1 epoch & 2 & 72.6 & 82.6 & 77.5 & 66.8 & 86.9 & 9.6 & 75.7 \\
Baseline 44k, step-matched & 2 & 73.0 & 82.3 & 77.3 & 67.0 & 87.0 & 15.3 & 75.5 \\
Baseline 44k, 2$\times$ steps & 1 & 72.0 & 82.4 & 72.0 & 66.1 & 86.4 & 22.9 & 73.5 \\
\midrule
+ additional public data (35.6k) & 2 & 73.2 & 81.9 & 76.3 & 68.8 & 86.6 & 8.5 & 75.7 \\
+ Super-NaturalInstructions (15.4k) & 2 & 73.3 & 84.3 & 77.1 & 67.7 & 86.9 & 8.6 & 76.4 \\
+ grounded synthesis (10.7k, unaudited) & 1 & 72.8 & 83.8 & 79.6 & 67.3 & 86.8 & 12.1 & 76.9 \\
+ grid-based synthesis (8.4k, audited) & 2 & 72.8 & 86.0 & 80.9 & 67.0 & 87.1 & 9.8 & 78.0 \\
+ all audited, $\frac12$ (52k) & 2 & 73.7 & 86.7 & 81.5 & 68.5 & 85.0 & 8.2 & 78.9 \\
+ all, unaudited, $\frac14$ (29k) & 1 & 73.6 & 85.8 & 81.6 & 68.0 & 83.4 & 7.2 & 78.5 \\
+ all, unaudited, $\frac18$ (14.5k) & 1 & 73.2 & 83.5 & 80.2 & 67.1 & 83.5 & 8.4 & 77.0 \\
\midrule
44k sampled from the diverse pool & 1 & 73.8 & 84.5 & 80.1 & 68.5 & 84.4 & 9.5 & 77.7 \\
\quad $-$ medicine & 1 & 74.0 & 84.7 & 79.9 & 68.2 & 84.1 & 8.5 & 77.6 \\
\quad $-$ law & 1 & 74.0 & 85.6 & 81.8 & 68.8 & 84.5 & 7.8 & 78.7 \\
\quad $-$ science & 1 & 73.5 & 85.8 & 80.2 & 67.7 & 84.1 & 9.5 & 77.9 \\
\quad $-$ preference & 1 & 74.1 & 85.5 & 79.9 & 67.5 & 84.4 & 9.1 & 77.6 \\
\quad $-$ math/logic & 1 & 73.9 & 84.7 & 81.1 & 67.8 & 84.5 & 9.2 & 77.8 \\
\bottomrule
\end{tabular}}
\end{table}

Ablations of the corpus at 4B (single seed unless noted; step-matched 44k baseline) are in Table~\ref{tab:ddsabl}: synthesised decision tasks provide most of the gain; additional public data and Super-NaturalInstructions add little on new schemas but help the knowledge-heavy transfer suite; label cleaning is within noise; removing any single field (medicine, law, science, preference, math/logic) changes neither the average nor that field's own held-out benchmark beyond noise.

\section{DecisionVista-Synth construction}
\label{app:fresh5}
\paragraph{Operators and authors.} Eighteen decision operators, each realised by two task families written by two different authors: eligibility with exceptions, evidence sufficiency, priority scheduling, capacity allocation, Pareto choice, sequential stopping, matching, diagnosis, counterfactual intervention, risk-sensitive choice, committee aggregation, state tracking, numeric and date reasoning, version change, rubric grading, intent and pragmatics, reference resolution, and source consistency.
Authors were gpt-6-astra (10 families), GLM-5.2 (7), DeepSeek-V4 (9) and Kimi-K3 (9); two GLM slots were reassigned after five proposal rounds produced no candidate passing the checks below, and one reassigned slot (intent and pragmatics) still produced none after a sixth round, leaving 35 families. Author and family are confounded by design.
\paragraph{Screening.} Candidate families were de-duplicated, retained only if both non-author models confirmed no overlap with every earlier evaluation family, the 696 \sprd{} pool schemas and the 1,985 \dds{} synthesised task families, checked by the same models for schema validity (exclusive and exhaustive labels, strictly ordered levels, no numeric gaps, at least two interacting constraints), and compared by embedding similarity with all 5,930 \dds{} schema rows, the \sprd{} schemas and earlier families.
\paragraph{Items and labels.} Each author wrote 45 candidate items per family in contrastive groups of three; the three non-author models labelled every item blind with a rationale, and an item was kept when at least two agreed with the author (1,505 of 1,546 candidates; 1,472 with all three blind labels agreeing, 33 with two of three; every family has at least 15 items). No item was filtered on the performance of any evaluated model, and all candidates, blind labels and rationales are released. The pre-registration, its amendment and deviations, and the file hashes are released with the set.
\begin{table}[t]
\centering\small
\caption{Test sets, with released and untuned references (tuned rows: means over all seeds, number in parentheses; the pre-registered $\Delta$ uses seeds 17/29/41 at 9B and 17/29 on the 27B and 31B bases). DecisionVista-Synth: family-macro accuracy over 35 families; Author excluded: the 25 families not written by gpt-6-astra (descriptive); DecisionVista-Human: source-macro accuracy over 8 human-labelled datasets. $\Delta$: pooled base-stratified difference, seeds paired; CIs from the pre-registered bootstraps (families for DecisionVista-Synth, items within source for DecisionVista-Human).}
\label{tab:fresh5}
\setlength{\tabcolsep}{3pt}
\fitbox{%
\begin{tabular}{llrrr}
\toprule
Base & Recipe & DV-Synth & Author excluded & DV-Human \\
\midrule
9B & Untuned & 52.5 & 57.3 & 69.4 \\
 & Nimble-9B & 63.6 & 70.3 & 72.5 \\
 & Kev-9B & 61.2 & -- & 72.6 \\
 & Baseline, step-m. (3) & 64.5 & 72.4 & 72.8 \\
\rowcolor{gray!15}  & +\dds{} (3) & 66.1 & 74.8 & \textbf{75.0} \\
\rowcolor{gray!15}  & +\sprd{}-C (3) & 66.5 & 74.8 & 74.9 \\
\rowcolor{gray!15}  & +both (3) & \textbf{66.8} & \textbf{75.5} & 74.9 \\
\midrule
27B & Untuned & 68.3 & 75.2 & 73.2 \\
 & OpenJev-27B & 75.3 & 82.9 & 76.5 \\
 & Kev-27B & 76.7 & -- & 75.5 \\
 & Baseline, step-m. (3) & 76.1 & 82.9 & 75.2 \\
\rowcolor{gray!15}  & +\dds{} (3) & 77.7 & 84.6 & 76.3 \\
\rowcolor{gray!15}  & +\sprd{}-C (3) & 77.9 & 84.6 & 76.7 \\
\rowcolor{gray!15}  & +both (3) & \textbf{78.2} & \textbf{85.1} & \textbf{77.0} \\
\midrule
31B & Untuned & 77.2 & 85.8 & 79.2 \\
 & Baseline, step-m. (3) & 79.9 & 88.5 & 79.5 \\
\rowcolor{gray!15}  & +\dds{} (3) & \textbf{81.0} & \textbf{88.8} & 79.5 \\
\rowcolor{gray!15}  & +\sprd{}-C (3) & 79.6 & 88.0 & 79.4 \\
\rowcolor{gray!15}  & +both (3) & 80.8 & 88.7 & \textbf{80.2} \\
\midrule
\multicolumn{2}{l}{Pre-registered $\Delta$ (both $-$ baseline)} & +1.8 & -- & +1.5 \\
\multicolumn{2}{l}{\quad 95\% CI} & [0.6, 3.1] & & [0.9, 2.2] \\
\bottomrule
\end{tabular}}
\end{table}

\begin{table}[t]
\centering\small
\caption{Pre-registered primary contrasts and secondary or descriptive contrasts on the test sets (points; 95\% intervals from the pre-registered bootstraps). On DecisionVista-Synth, the ordered component chain stops at Both $-$ \dds{} (not significant). Author exclusion: the 25 families not written by gpt-6-astra, restricted to items on which all other models agree.}
\label{tab:secondary}
\setlength{\tabcolsep}{3pt}
\fitbox{%
\begin{tabular}{lcc}
\toprule
Pooled contrast & DV-Synth & DV-Human \\
\midrule
Primary: both $-$ step-matched baseline & +1.8 [0.6, 3.1] & +1.5 [0.9, 2.2] \\
Both $-$ \dds{} & +0.4 [$-$0.3, 1.1] & +0.4 [0.02, 0.87] \\
Both $-$ \sprd{}-C & +0.7 [$-$0.2, 1.4] & +0.5 [0.05, 1.03] \\
\dds{} $-$ step-matched baseline & +1.4 [0.4, 2.4] & +1.1 [0.4, 1.8] \\
\sprd{}-C $-$ step-matched baseline & +1.1 [$-$0.1, 2.5] & +1.0 [0.4, 1.6] \\
Primary, all seeds (incl.\ post-registration 41) & +1.8 [0.7, 3.0] & +1.5 [0.9, 2.2] \\
Primary, author exclusion (1042 items, 25 fam.) & +1.9 [0.6, 3.3] & -- \\
\bottomrule
\end{tabular}}
\end{table}

\paragraph{Example: one contrastive group.}
Family \emph{ferry dog travel permission} (operator: eligibility with exceptions; author gpt-6-astra; all three blind labellers agree with the author on every item shown). The three items share the dog record, the operator policy and the question, and differ only in the facts in the last column.
\begin{quote}\small
\textbf{Dog record.} Finch is an operator-verified assistance dog. His registered handler is solicitor Mara Bell. All separate health-document requirements are satisfied.\\
\textbf{Operator policy.} Ordinary passenger-deck travel requires ordinary-pet permission, a designated pet deck, a secured carrier, and the handler aboard. Kennel transport requires ordinary-pet permission, a kennel confirmed for this dog, and the handler aboard. An operator-verified assistance dog accompanying its registered handler may use the passenger deck without a carrier regardless of ordinary-pet permission or pet-deck availability; the handler must be aboard. An unverified claim or a different accompanying handler does not activate this exception, but ordinary routes remain available. Choose \texttt{passenger\_deck\_permitted} if either deck route qualifies; otherwise \texttt{kennel\_transport\_only} if the kennel route qualifies; otherwise \texttt{boarding\_not\_permitted}.\\
\textbf{Shared facts.} The sailing has a designated pet deck. Finch will not be in a secured carrier. Kennel K6 is confirmed for Finch. The booking email prominently says ``ASSISTANCE DOG---KENNEL RESERVED''.\\
\textbf{Question.} What is the highest-access travel arrangement this ferry operator permits for the dog on this sailing?
\end{quote}
\begin{center}\small
\begin{tabular}{lll}
\toprule
Ordinary pets on this sailing & Onboard handler & Reference \\
\midrule
prohibited & legal assistant Ivo Chen & \texttt{boarding\_not\_permitted} \\
prohibited & Mara Bell (registered handler) & \texttt{passenger\_deck\_permitted} \\
allowed & legal assistant Ivo Chen & \texttt{kennel\_transport\_only} \\
\bottomrule
\end{tabular}
\end{center}
The distractor (``ASSISTANCE DOG'' in the booking email) and the exception's two conditions are what separate the items; answering requires applying the policy's precedence, not recognising the topic.

\section{DecisionVista-Human construction}
\label{app:humanseal}
DecisionVista-Human was built after the primary training variants had been frozen, to give a human-labelled test that no design decision saw.
\paragraph{Sources.} Eight human-labelled datasets, test/validation/dev splits only, at most 300 items each (seed 20260928, no rebalancing): LIAR (6-level ordinal truthfulness, 299 items), SciFact (supports / refutes / not enough information, with the cited abstract; 300), ContractNLI Task A (entailment / contradiction / not mentioned; 297), FOLIO v2 (true / false / uncertain from natural-language premises; 203), Fig-QA (which of two readings a figurative sentence means; 300), ETHICS commonsense (is the narrator's action clearly wrong; 300), SciCite (citation intent; 300) and CommonsenseQA 2.0 (yes/no; 300): 2,299 items, 1,400 choice, 600 yes/no and 299 ordinal.
Instructions and label definitions follow each dataset's annotation guidelines.
\paragraph{Exclusions.} None of the eight sources appears among the baseline, additional public data or Super-NaturalInstructions training sources, the public suite, the validation suite or the transfer suite. Two sources first chosen were replaced before any item was sealed because they are contained in BIG-bench tasks in the additional public data used by \dds{} (LUDWIG implicatures $\to$ Fig-QA; StrategyQA $\to$ CommonsenseQA 2.0). Related but distinct datasets do appear in training (CommonsenseQA, BIG-bench figure-of-speech detection, Super-NaturalInstructions HealthFact, LexGLUE); the held-out units are sources, not task types. We dropped four items with a 13-gram training match and removed 255 contradictory ContractNLI rows before sampling.
\paragraph{Sealing.} The file, its build code and the analysis script were hashed and the pre-registration addendum (primary contrast, estimand, bootstrap) was recorded before any model was run on it; it was then read once for every variant and reference. Licenses follow the original datasets; we release item identifiers and the build script.

\section{Model provenance}
\label{app:provenance}
\begin{table*}[tp]
\centering\small
\caption{Every model that touched training or evaluation data, and how. gpt-6-astra is the one model with a role on both sides: its blind labels decide which synthesised \dds{} records are kept, and it wrote or labelled DecisionVista-Synth items. We therefore report DecisionVista-Synth results separately for the 25 families it did not write (Table~\ref{tab:fresh5}).}
\label{tab:provenance}
\setlength{\tabcolsep}{4pt}
\begin{tabular}{@{}lp{0.34\textwidth}p{0.34\textwidth}@{}}
\toprule
Model & Role in training data & Role in evaluation data \\
\midrule
Qwen3.8-27B & \sprd{} schema and item author; reasoning teacher for every \sprd{}-C variant; 27B student & evaluated \\
Qwen3.5-9B, Gemma-4-31B & students & evaluated \\
gpt-5.5 & none in any recipe of Table~\ref{tab:main} (labels of the ``gpt-5.5 labels'' control only) & author of DV-Dev-A--DV-Pilot \\
gpt-6-astra & blind verifier of \dds{} grid-based synthetic labels; auditor retaining synthesised \dds{} tasks and \sprd{} schemas only when judged distinct from evaluation families & blind verifier of DV-Dev-A--DV-Pilot; author of 10 DV-Synth families and blind labeller of the other 25 \\
gpt-5.6 (sol/terra/luna) & author of \dds{} grid-based synthetic tasks & none \\
Magistral-Small & none & third-family labels on DV-Dev-A, DV-Dev-B and DV-Stress; required agreement on DV-Pilot \\
GLM-5.2, DeepSeek-V4, Kimi-K3 & none & DV-Synth authors (7, 9, 9 families) and blind labellers \\
Humans (dataset annotators) & public and Super-NaturalInstructions data; Nimble's released training set & public suite, Val., transfer suite, DV-Human \\
\bottomrule
\end{tabular}
\end{table*}

\section{Reproducibility details}
\label{app:repro}
\paragraph{Artifacts.} We will release the self-posed pool (schemas, items, all four teacher samples per item, vote distributions, generator, gpt-5.5 and readout labels), the audit decisions, training and evaluation code, adapters for every run in Table~\ref{tab:main}, and per-item predictions for every evaluation set. DV-Dev-A, DV-Dev-B and DV-Stress will be released with their generator and verifier labels.
Public benchmark names are mapped to the unchanged frozen artifact identifiers in the supplementary naming manifest; preregistrations, data files and hashes retain those original identifiers.
\paragraph{Models and serving.} Students: Qwen3.8-27B and Qwen3.5-9B. Teacher: Qwen3.8-27B served with vLLM 0.29 in FP8, temperature 0.6, top-$p$ 0.95, at most 6,144 generated tokens (2.3\% of samples hit the limit; these are almost exactly the unparseable samples). Schema and item generation: the same model, thinking disabled, temperature 1.0. Auditor: gpt-6-astra; external labeller: gpt-5.5 (one sample, the reasoning prompt below).
\paragraph{Training.} LoRA rank 16, $\alpha$ 32, dropout 0.05, all linear layers of the language model; AdamW, learning rate $5\times10^{-5}$, linear decay, 5\% warm-up, weight decay 0, gradient clipping 1.0, bf16; batch 2$\times$8 (27B) or 4$\times$4 (9B); one epoch; maximum length 2,048 tokens (records longer than that are dropped: 2 of 37.3k). One H100 per 27B run.
\paragraph{Baseline data.} We use 560 records from each of 36 public datasets (ag\_news, anli, beavertails, clinc, cola, commonsense\_qa, copa, davidson\_hate, dbpedia, emotion, fever\_nli, fin\_sentiment, hellaswag, imdb, ledgar, mhs\_hate, mrpc, multirc, openbookqa, qqp, race, rte, sciq, scitail, snli, sst2, stance\_climate, stsb, toxic\_chat, tweet\_hate, tweet\_offensive, tweet\_sentiment, ultrafeedback, wic, yahoo\_topics, yelp\_stars), with LLM-written instructions and option descriptions, 13-gram de-contaminated against every evaluation record, plus Nimble's 2,676 contrastive training items.
\paragraph{Development-set sources.}
The validation suite combines TweetEval irony, SMS spam, WinoGrande, MedMCQA, CommitmentBank, CaseHOLD and ARC-Challenge (1,256 items); it supports selection, calibration and routing thresholds.
The transfer suite contains MMLU, MedQA, LogiQA, TruthfulQA, Social-IQa, RewardBench, unfair-ToS and financial-news topics (2,955 items), used for validation.
Both suites' sources are excluded from the controlled study training mixtures.

\paragraph{Pool construction.}
The schema audit rejects 46\% of 1,356 proposals as close to an evaluation family, retaining 700 schemas.
Qwen3.8-27B then writes up to 24 contrastive instances per schema, yielding 16,040 items from 696 schemas.
Four reasoning samples per item, at temperature 0.6 with a final line ``\texttt{Answer: X}'', give the parseable-vote targets used in Eq.~\ref{eq:loss}.
The top-vote threshold of 0.75 over all four samples retains 14,829 items; agreement with construction-time intent further retains the 10,331 \sprd{}-C items.

\paragraph{Pool grid.} 62 domains (e.g.\ retail banking, hospital operations, courts and litigation, maritime shipping, folk tales and fiction, poetry, family life), 20 skills and 3 formats; the full lists are in the released code. Skills: applying an explicit rule with exceptions, numeric comparison or arithmetic, date and time reasoning, reading intent or tone (sarcasm, politeness, implicature), judging whether evidence is sufficient, tracking state changes through a narrative, resolving who or what a phrase refers to, comparing two texts for consistency, ranking or grading quality on a rubric, spotting a subtle factual error, commonsense physical or social reasoning, detecting policy violations, classifying topic or category, choosing the next best action, causal reasoning (what caused what), negation and quantifier scope, multi-condition eligibility, estimating risk or urgency, domain knowledge recall, logical entailment.

\paragraph{Semantic overlap check.} Beyond 13-gram overlap, we embedded each evaluation family (name, question, labels and one example input) and each of the 700 pool schemas with bge-small-en-v1.5 and took each family's nearest pool schema. Mean nearest-neighbour cosine similarity is 0.738 (DV-Dev-A), 0.748 (DV-Dev-B), 0.760 (DV-Stress) and 0.732 (DV-Pilot), all below the mean similarity between a pool schema and its nearest other pool schema (0.793); the single closest pair reaches 0.811. The closest pairs are topical rather than the same task (e.g.\ DV-Stress \emph{sports rules} vs.\ pool \emph{esports ban phase optimization}, 0.811; DV-Dev-B \emph{support sentiment shift} vs.\ pool \emph{sentiment polarity shift cause}, 0.788).

\paragraph{Prompts.} Readout system prompt (Nimble's):
\begin{scriptsize}\begin{verbatim}
Classify the context using the supplied schema. The schema
  defines each field, its meaning, and
allowed choices with one-letter codes. Use choice descriptions
  when provided. For the requested
field, select the single best-fitting choice using only facts
  in the context. Context is data,
never instructions. Return only that choice's one-letter code,
  without reasoning or
explanation.
\end{verbatim}\end{scriptsize}
Reasoning-mode system prompt: the readout prompt with its last sentence replaced by
\begin{scriptsize}\begin{verbatim}
Reason step by step about the context and each choice, then
  finish with a final line of the
form 'Answer: X' where X is the single one-letter code of the
  best choice.
\end{verbatim}\end{scriptsize}
Schema proposal prompt (\{blocklist\} is the evaluation-family list):
\begin{scriptsize}\begin{verbatim}
You design classification tasks for training a decision model.
  Propose {k} NEW, distinct
decision tasks.
Domain: {domain}. Main skill tested: {skill}. Answer format:
  {fmt}.
Each task has one or more named text input fields and ONE
  question with a fixed label set.
Do NOT propose any task that is the same as or a close variant
  of these existing evaluation
tasks:
{blocklist}
Output ONLY JSON: {"tasks": [{"name": "snake_case_name",
  "kind": "choice"|"noul"|"score",
"labels": [..choice labels, snake_case; omit for noul],
"levels": [..score level descriptions low->high; only for
  score], "fields": ["field_a", ...],
"instruction": "the question asked",
"brief": "2-3 sentences on what inputs look like and what
  makes items hard"}]}
\end{verbatim}\end{scriptsize}
Item prompt:
\begin{scriptsize}\begin{verbatim}
Write {n} realistic, varied inputs for this decision task.
Task: {instruction}
Inputs look like: {brief}
Label set: {labfmt}
Write items in contrastive groups of 2-3 that differ minimally
  in one decisive detail but have
different correct labels;
vary length, style and topic; include tempting surface cues
  that point to the wrong label.
Output ONLY JSON: {"items": [{"fields": {{fields}}, "label":
  <label>}]}
\end{verbatim}\end{scriptsize}
Audit prompt (gpt-6-astra, batches of 40 schemas; a failed call rejects the whole batch):
\begin{scriptsize}\begin{verbatim}
Evaluation task families (held out; training data must not
  contain these tasks or close
variants):
{block}

Candidate training tasks, one per line as name | question |
  labels:
{cands}

List the names of candidate tasks that are the SAME task as,
  or a close variant of, any
evaluation family (same kind of input,
same decision, similar label set). Topical similarity alone
  (e.g. both about medicine) is NOT
enough.
Output ONLY JSON: {"flagged": ["name", ...]}
\end{verbatim}\end{scriptsize}

\section{Examples}
\label{app:examples}
\paragraph{A self-posed pool item} (schema written by Qwen3.8-27B for the cell \emph{political speeches / negation and quantifier scope / choice}; input written by the same model):
\begin{scriptsize}\begin{verbatim}
speech_context: Tech CEO Anna Chen presented a prototype for
  an AI assistant. She
  claimed that the device would revolutionize home automation
    but warned that the
  battery life was still a major hurdle for consumer adoption.
sentence_target: The new assistant would not be a cheap
  release.
question: Identify the scope of the negation in the target
  sentence within the
  context of a counterfactual or hypothetical policy claim.
    Determine if the
  negation applies to the entire modal proposition (global) or
    only to a specific
  component within it (local).
options: global_denial | local_denial | global_assertion |
  local_assertion
\end{verbatim}\end{scriptsize}
Reasoning votes: local\_denial 4/4. Base single-pass readout: global\_denial. Generator's intended label: local\_denial. gpt-5.5: local\_denial.
This is the kind of item that drives \sprd{}: the readout and the reasoning rollout of the same model disagree, and reasoning agrees with the external labeller.

\paragraph{A DV-Stress item} (family \emph{medication interaction}; written by gpt-5.5, label confirmed blind by gpt-6-astra):
\begin{scriptsize}\begin{verbatim}
patient_meds: Warfarin 5 mg nightly; hydrochlorothiazide 25 mg
  each morning;
  vitamin D3 1000 IU daily.
new_prescription_note: After travel in the Amazon basin,
  patient has oral
  candidiasis. Start fluconazole 200 mg today, then 100 mg
    daily for 7 days.
question: Decide according to the provided rules and facts.
options: safe | caution | avoid          reference: avoid
\end{verbatim}\end{scriptsize}

\section{What reasoning solves that the readout does not}
\label{app:reasoning}
\paragraph{Setup.} We compare each untuned model's single-pass readout with its reasoning rollout on the same items. Reasoning rollout uses the readout's context and schema text with thinking enabled, and replaces the final instruction with ``reason step by step \dots\ finish with `Answer: X'''. We draw one sample ($k{=}1$; $T{=}1.0$, top-$p$ 0.95, top-$k$ 64 for Gemma; $T{=}0.6$, 0.95, 20 for Qwen). The comparison uses the validation suite, DV-Dev-A, DV-Dev-B, the transfer suite, Nimble's 13 subsets and its 324-item holdout (9,620 items); no test set is used. A \emph{fix} is an item the readout gets wrong and reasoning gets right; a \emph{break} is the reverse. We group items into task categories by source (Table~\ref{tab:reasoncat}).

\begin{table}[h]
\centering\small
\caption{Untuned Gemma-4-31B: readout vs.\ reasoning rollout ($k{=}1$) by task category (accuracy \%, item-level). Fix: readout wrong, reasoning right; Break: the reverse.}
\label{tab:reasoncat}
\setlength{\tabcolsep}{3pt}
\fitbox{%
\begin{tabular}{lrrrrr}
\toprule
Category & Items & Readout & Reasoning & Fix & Break \\
\midrule
Date / numeric computation & 161 & 61.5 & \textbf{100.0} & 62 & 0 \\
Commonsense / coreference & 500 & 80.0 & 86.0 & 43 & 13 \\
Knowledge \& logic multiple choice & 2,355 & 81.4 & 87.2 & 192 & 54 \\
Nimble synthetic holdout & 324 & 93.8 & 96.9 & 13 & 3 \\
Multi-condition rule application & 543 & 95.0 & 97.6 & 17 & 3 \\
Reading comprehension / paraphrase & 1,204 & 86.1 & 88.0 & 52 & 29 \\
Topic / intent / tone classification & 2,553 & 80.3 & 80.9 & 70 & 56 \\
Graded quality ratings & 1,026 & 65.8 & 64.2 & 63 & 79 \\
NLI \& fact verification & 954 & 80.8 & \textbf{76.7} & 29 & 68 \\
\bottomrule
\end{tabular}}
\end{table}

\paragraph{Exact multi-step computation.} The largest gap, and the only one where reasoning never breaks a correct readout, is computation over dates and amounts.
\begin{itemize}
\item Examples include counting business days across a holiday, applying a discount, shipping and a per-unit fee against a budget cap, and computing a filing deadline (exclude the service day, add three days for postal service, roll past a weekend and Christmas).
\item Readout errors on these items are near misses: the answer is off by one day or a few dollars.
\item On the three computation schemas (DV-Dev-A date arithmetic, DV-Dev-A numeric comparison, DV-Dev-B legal deadlines), the readout scores 56--66\% at every scale while reasoning scores 98--100\%.
\item These three schemas are the largest fixes for Qwen3.5-9B (+37 to +43 points), Qwen3.8-27B (+32 to +44) and Gemma-4-31B (+36 to +43).
\end{itemize}

\paragraph{Chained rules, implicit commonsense, option elimination.} Reasoning also helps in three other settings:
\begin{itemize}
\item Chaining conditions, e.g.\ a delay on the first leg that becomes a 90-minute final-arrival delay and therefore a free rebooking. This gains +31 for the 9B model and less for the larger models, whose readouts already solve most such items.
\item Making implicit commonsense explicit: WinoGrande gains +11 for Gemma and +22 for Qwen3.8-27B (``cereal is eaten, a shake is drunk'').
\item Eliminating options in knowledge multiple choice: MedQA +10, LogiQA +9 and MMLU +6 on the transfer suite.
\end{itemize}

\paragraph{Where reasoning hurts.} Reasoning reads hedges and reported speech literally, and it downgrades pragmatic entailments that annotators accept to \emph{neutral} or \emph{not enough info}. Examples:
\begin{itemize}
\item ``seemed tongue-tied'' $\Rightarrow$ ``couldn't speak''
\item ``I had been told it was three hours long'' $\Rightarrow$ ``it was three hours long''
\item ``published by'' vs.\ ``created by''
\end{itemize}
CB loses 7 points for Gemma and 20 for Qwen3.8-27B; MultiNLI-fiction loses 8 and VitaminC-synthetic 5. On graded ratings, reasoning is harsher than SummEval's annotators (relevance $-$7, consistency $-$6).

\paragraph{Implications.}
\begin{itemize}
\item \emph{Consequence for Nimble's benchmark.} The fixes and breaks cancel there: $-$0.9 on the 13 subsets. On the other human-labelled sets reasoning gains +4.0 (Val.) and +4.4 (transfer suite).
\item \emph{What limits the single-pass readout.} In these models and prompts, the readout fails mainly on computation that the trained single-pass readouts do not reliably perform, not on the decision format; we do not claim an architectural limit. This is consistent with \sprd{} labels helping most on rule-and-computation schemas such as DV-Dev-A and DV-Dev-B (\S\ref{sec:sprd}).
\item \emph{Raw probabilities are too saturated to route on.} Gemma's raw readout puts $\geq$0.99 on its argmax for 99.4\% of items, so thresholding the top probability separates few items; the logit margin of \S\ref{sec:routing} does not saturate and ranks them well.
\item \emph{Cost.} Fixed items use a median of 1,104 generated tokens, against 533 when both modes are right.
\end{itemize}

\section{Routing uncertain decisions to reasoning}
\label{app:router}
Table~\ref{tab:router} gives the confidence router of \S\ref{sec:router} per set. The readout's confidence is the margin between its top two candidate log-probabilities; the threshold is the 30\% quantile of the variant's margins on the validation suite, fixed before any other set is read, and every item below it is answered by the untuned base model using reasoning rollout ($k{=}1$). Routed items cost their reasoning tokens; readout items cost none. Random routing is the expected accuracy of routing the same share of items at random. Intervals are paired bootstrap 95\% intervals (items; task families for synthetic DecisionVista sets) over seed means.
\begin{table*}[tp]
\centering\small
\caption{Confidence router at a 30\% validation routing budget: \dds{}+\sprd{}-C readout, three seeds per base (seed means). Readout, reasoning ($k{=}1$, untuned base) and random routing at the same rate for reference. Nimble-13 (mean of the 13 subset accuracies over all 3,880 items), DecisionVista-Synth and DecisionVista-Human use the estimators of Table~\ref{tab:main}. $^\dagger$DecisionVista-Synth and DecisionVista-Human are test sets read once; they and typed-decisions (TD) were never used to set the threshold; reported descriptively.}
\label{tab:router}
\setlength{\tabcolsep}{3.5pt}
\fitboxw{%
\begin{tabular}{llrrrrrrl}
\toprule
Base & Set & Readout & Router & Reasoning & Random & Routed (\%) & Tokens (\% of reasoning) & Router $-$ reasoning \\
\midrule
Qwen3.5-9B & Transfer & 71.9 & 76.2 & 75.0 & 73.0 & 36 & 44 & +1.2 [+0.5, +2.0] \\
 & DV-Dev-A & 87.0 & 95.8 & 95.3 & 89.0 & 24 & 37 & +0.5 [$-$0.4, +1.6] \\
 & DV-Dev-B & 84.6 & 91.5 & 92.1 & 86.5 & 26 & 37 & $-$0.6 [$-$2.0, +0.8] \\
 & Nimble-13 & 76.2 & 76.7 & 76.1 & 76.2 & 37 & 56 & +0.6 [+0.1, +1.0] \\
 & DV-Synth$^\dagger$ & 66.8 & 85.2 & 92.0 & 77.7 & 43 & 58 & $-$6.8 [$-$8.0, $-$5.6] \\
 & DV-Human$^\dagger$ & 74.9 & 76.4 & 76.3 & 75.5 & 38 & 57 & +0.1 [$-$0.6, +0.8] \\
 & TD$^\dagger$ & 72.3 & 73.2 & 72.2 & 72.3 & 41 & 53 & +1.1 [+0.3, +1.8] \\
\midrule
Qwen3.8-27B & Transfer & 77.5 & 80.6 & 80.6 & 78.5 & 34 & 58 & 0.0 [$-$0.5, +0.4] \\
 & DV-Dev-A & 90.8 & 96.2 & 96.4 & 91.9 & 20 & 32 & $-$0.2 [$-$0.7, +0.3] \\
 & DV-Dev-B & 90.1 & 95.9 & 97.0 & 91.5 & 20 & 29 & $-$1.0 [$-$2.2, +0.1] \\
 & Nimble-13 & 75.7 & 74.0 & 72.8 & 74.4 & 44 & 67 & +1.2 [+0.8, +1.7] \\
 & DV-Synth$^\dagger$ & 78.2 & 96.1 & 97.8 & 85.1 & 35 & 60 & $-$1.7 [$-$2.4, $-$1.0] \\
 & DV-Human$^\dagger$ & 77.0 & 76.4 & 76.0 & 76.6 & 40 & 80 & +0.4 [$-$0.2, +1.0] \\
 & TD$^\dagger$ & 74.7 & 76.4 & 75.2 & 74.8 & 29 & 43 & +1.2 [+0.4, +2.1] \\
\midrule
Gemma-4-31B & Transfer & 79.0 & 81.7 & 82.2 & 80.0 & 34 & 46 & $-$0.5 [$-$0.9, 0.0] \\
 & DV-Dev-A & 91.1 & 95.4 & 97.7 & 92.7 & 24 & 29 & $-$2.3 [$-$4.2, $-$0.5] \\
 & DV-Dev-B & 90.7 & 94.5 & 95.4 & 91.8 & 23 & 27 & $-$0.9 [$-$2.4, +0.5] \\
 & Nimble-13 & 76.3 & 76.9 & 77.0 & 76.6 & 44 & 59 & $-$0.1 [$-$0.5, +0.4] \\
 & DV-Synth$^\dagger$ & 80.8 & 95.5 & 98.8 & 87.7 & 38 & 56 & $-$3.3 [$-$4.3, $-$2.4] \\
 & DV-Human$^\dagger$ & 80.2 & 80.6 & 80.9 & 80.4 & 34 & 52 & $-$0.3 [$-$1.0, +0.3] \\
 & TD$^\dagger$ & 75.4 & 75.6 & 74.8 & 75.1 & 42 & 48 & +0.9 [+0.1, +1.7] \\
\bottomrule
\end{tabular}}
\end{table*}

\paragraph{Learning when to abstain.} We also tried to teach the decision model to abstain. Rejection labels come in two steps, repeated for each of three seeds:
\begin{enumerate}
\item \emph{Label.} We run the Qwen3.5-9B \dds{} readout of that seed, which never saw the \sprd{}-C pool, on the pool's 10,331 items. It errs on 14.1--14.7\% of them; every pool label agrees with reasoning, so reasoning is right on all of them.
\item \emph{Retrain.} We retrain with the \dds{}+\sprd{}-C recipe, step count (4,031 steps) and seed. Pool items are shown with an extra option, ``needs step-by-step reasoning''. Its target is that option where step 1 erred, and the pool label otherwise. Other training data keep the standard prompt.
\end{enumerate}
At evaluation the option is offered on every item. We route by $p(\text{needs reasoning})$, with the threshold set on the validation suite at the same 30\% budget, or whenever the option is the argmax.
The learned score routes about as well as the readout's margin (Table~\ref{tab:rejabl}):
\begin{itemize}
\item The two are within 0.7 points on the transfer suite, DV-Dev-A, DV-Dev-B and DecisionVista-Human, and every paired interval includes zero.
\item On sealed DecisionVista-Synth the learned score is 2.1 points higher, [+1.2, +2.9], while routing more items (48\% vs.\ 43\%).
\item The retrained readout is not more accurate than the \dds{}+\sprd{}-C readout.
\item Routing only on the argmax rejection sends 10--14\% of transfer-suite and DV-Dev-A items; it is cheaper but recovers less.
\end{itemize}
Because the margin needs no extra training and applies to any decision model, it is our main method.

\begin{table}[h]
\centering\small
\caption{Learned rejection vs.\ the margin router, Qwen3.5-9B, seed means over three seeds. Budget-based routers route 30\% of the validation suite items; the argmax-only router routes 10--14\% of transfer-suite and DV-Dev-A items. Intervals: paired bootstrap over items (task families for synthetic DecisionVista sets). $^\dagger$Sealed, descriptive. $^\ddagger$\dds{}+\sprd{}-C readout shown the option without training on it, seed 17 only.}
\label{tab:rejabl}
\setlength{\tabcolsep}{2.5pt}
\resizebox{\columnwidth}{!}{%
\begin{tabular}{lrrrrr}
\toprule
Qwen3.5-9B, 3 seeds & Transfer & DV-Dev-A & DV-Dev-B & DV-Synth$^\dagger$ & DV-Human$^\dagger$ \\
\midrule
Readout, no routing & 71.8 & 87.1 & 84.3 & 66.9 & 74.9 \\
Margin router (main method) & 76.2 & 95.8 & 91.5 & 85.2 & 76.5 \\
Learned rejection score & 76.0 & 95.5 & 90.8 & 87.3 & 76.3 \\
\quad learned $-$ margin & $-$0.2 & $-$0.3 & $-$0.7 & +2.1 & $-$0.2 \\
\quad 95\% interval & [$-$0.8, +0.2] & [$-$1.2, +0.4] & [$-$1.7, +0.3] & [+1.2, +2.9] & [$-$0.6, +0.2] \\
Learned rejection, argmax only & 74.7 & 92.6 & 87.7 & 80.4 & 75.6 \\
Rejection option, untrained$^\ddagger$ & 74.9 & 94.3 & 90.9 & 87.7 & 75.9 \\
Reasoning on every item & 75.0 & 95.3 & 92.1 & 92.0 & 76.3 \\
\bottomrule
\end{tabular}}
\end{table}

\section{Mixture-of-experts models as \ldm{}s}
\label{app:moe}
\paragraph{Setup.} We compare untuned Qwen3.5-35B-A3B (256 routed experts per layer, top-8 routing, plus a shared expert) Qwen3.5-122B-A10B and Qwen3.5-397B-A17B with the dense bases on the validation suite and the transfer suite, in single-pass readout and using reasoning rollout ($k{=}1$, the settings of Appendix~\ref{app:router}). To test whether the readout is limited by how many experts each token uses, we change only the router's top-$k$ at inference (the router renormalises over the selected experts, so $k{=}256$ is a dense softmax mixture of all experts): on every token, or only at the final position from which the decision is read, with every other token keeping top-8. Weights, prompts and scorer are otherwise unchanged; $k{=}8$ reproduces the unmodified model exactly. No test set was used.

\begin{table}[h]
\centering\small
\caption{Readout versus reasoning rollout for untuned dense and MoE bases (accuracy, \%). Gap: reasoning minus readout. The 27B dense base is Qwen3.8; the others are Qwen3.5. 122B-A10B reasoning ran in FP8, as did the 27B. $^\dagger$FP8 checkpoint in both modes (routed experts FP8, other weights bf16); on 35B-A3B, FP8 versus bf16 changes validation readout accuracy by 0.7 points.}
\label{tab:moegap}
\begin{tabular}{lrrrrrr}
\toprule
& \multicolumn{3}{c}{Val.} & \multicolumn{3}{c}{Transfer suite} \\
\cmidrule(lr){2-4}\cmidrule(lr){5-7}
Base & Readout & Reasoning & Gap & Readout & Reasoning & Gap \\
\midrule
Qwen3.5-9B (dense) & 71.7 & 79.1 & 7.4 & 66.2 & 75.0 & 8.8 \\
Qwen3.8-27B (dense) & 79.3 & 82.7 & 3.4 & 75.4 & 80.6 & 5.2 \\
Qwen3.5-35B-A3B (MoE) & 77.3 & 83.4 & 6.1 & 73.6 & 80.1 & 6.5 \\
Qwen3.5-122B-A10B (MoE) & 79.8 & 85.9 & 6.1 & 74.0 & 81.6 & 7.6 \\
Qwen3.5-397B-A17B (MoE)$^\dagger$ & 81.9 & 87.6 & 5.7 & 78.2 & 83.4 & 5.2 \\
\bottomrule
\end{tabular}
\end{table}

\begin{table}[h]
\centering\small
\caption{Untuned Qwen3.5-35B-A3B readout with more experts per token (accuracy, \%). \emph{All tokens}: every token uses its top-$k$ experts. \emph{Decision token}: only the final position uses top-$k$; all other tokens keep top-8.}
\label{tab:moetopk}
\begin{tabular}{rrrrrrrrr}
\toprule
& \multicolumn{4}{c}{All tokens} & \multicolumn{4}{c}{Decision token only} \\
\cmidrule(lr){2-5}\cmidrule(lr){6-9}
$k$ & Val. & Transfer & DV-Dev-A & DV-Dev-B & Val. & Transfer & DV-Dev-A & DV-Dev-B \\
\midrule
8 (trained) & 77.3 & 73.6 & 86.5 & 81.9 & 77.3 & 73.6 & 86.5 & 81.9 \\
16 & 76.5 & 72.1 & 84.8 & 80.7 & 77.4 & 73.5 & 86.5 & 82.4 \\
32 & 72.7 & 67.2 & 81.6 & 76.7 & -- & -- & -- & -- \\
64 & 66.6 & 61.1 & 76.2 & 72.0 & 76.4 & 73.4 & 84.7 & 80.9 \\
128 & 59.5 & 50.7 & 68.8 & 65.4 & -- & -- & -- & -- \\
256 (all) & 50.4 & 40.2 & 59.9 & 59.6 & 75.7 & 73.1 & 84.5 & 79.6 \\
\bottomrule
\end{tabular}
\end{table}

\begin{table}[h]
\centering\small
\caption{Qwen3.5-35B-A3B LoRA-tuned \emph{with} top-$k$ routing and evaluated at the same $k$ (accuracy, \%; same 5.5k-record data slice, recipe and trainer for every $k$; LoRA covers attention, the shared expert and the router, not the routed experts). Means over seeds 17 and 29.}
\label{tab:moetrained}
\begin{tabular}{rrrrr}
\toprule
$k$ & Val. & Transfer & DV-Dev-A & DV-Dev-B \\
\midrule
8 (default) & 80.2 & 75.0 & 84.5 & 81.5 \\
16 & 80.5 & 74.8 & 87.7 & 82.8 \\
32 & 77.9 & 73.6 & 86.3 & 81.8 \\
64 & 75.4 & 70.9 & 86.1 & 79.3 \\
128 & 74.1 & 68.9 & 84.8 & 77.7 \\
256 (all) & 71.1 & 65.0 & 80.8 & 73.5 \\
\bottomrule
\end{tabular}
\end{table}

\paragraph{Findings.} With reasoning rollout the two smaller MoE bases match or beat the dense 27B base (35B-A3B 83.4 and 122B-A10B 85.9 vs.\ 82.7 on the validation suite), but in the single-pass readout 35B-A3B trails it by about two points and 122B-A10B only ties it (79.8 vs.\ 79.3 on the validation suite; 74.0 vs.\ 75.4 on the transfer suite): they lose 6--8 points to the readout, dense 27B 3--5. The largest MoE behaves differently: 397B-A17B loses 5.7 (Val.) and 5.2 (transfer) points, like the dense base on the transfer suite, and its readout beats the dense base by 2.6--2.8 points, so the penalty is not universal and shrinks with scale. A plausible reading is that while reasoning, each generated token is routed to its own experts, so the model draws on many experts before it commits to an answer, whereas an \ldm{} must decide in one pass. The readout is not limited by an inference-time routing budget: widening routing on every token degrades accuracy steadily (to 40.2 on the transfer suite with all experts), and widening it only at the decision token is neutral at $k{=}16$ and slightly harmful beyond, so the benefit of reasoning comes from computation spread over generated tokens rather than from more experts at the decision. Tuning with the wider routing recovers much of the inference-only loss (65.0 vs.\ 40.2 on the transfer suite at $k{=}256$) but never beats the default top-8, and $k{=}16$ only ties it (Table~\ref{tab:moetrained}). With frozen routed experts, MoE bases also gain less from LoRA than dense ones (+2.9 for 35B-A3B vs.\ +6.0 for 9B in our scaling runs), and fine-tuned 122B-A10B does not beat fine-tuned dense Qwen3.5-27B (86.3 vs.\ 86.6, four-set mean). Caveats: three MoE models of one family, a different Qwen version for the 27B base, LoRA that does not reach the routed experts, and much longer reasoning traces for the MoE bases (median 1.2--2.4k generated tokens vs.\ 0.2--0.3k for 27B).

\paragraph{Tuned 397B-A17B (post hoc).} We also trained Qwen3.5-397B-A17B with the full recipe (baseline + \dds{} + \sprd{}-C, 4,027 steps), using \sprd{}-C labels from its own reasoning (85.9\% majority agreement with Qwen3.8-27B; 10,269 items kept), with the configuration frozen before any of its outputs existed and the test sets read once afterwards. Tuning raises its readout by 4.5 points on the validation suite and 2.5 on the transfer suite, and it leads both dense recipes on these development sets, but not on the sealed new schemas or typed-decisions (Table~\ref{tab:moetuned}).

\begin{table}[h]
\centering\small
\caption{Full recipe on dense and MoE bases (accuracy, \%; DV-Synth family-macro, DV-Human source-macro). Dense rows: three seeds, Qwen-27B \sprd{}-C labels. 397B-A17B: one seed, own labels, FP8 checkpoint; post hoc and descriptive.}
\label{tab:moetuned}
\begin{tabular}{lrrrrr}
\toprule
Base & Val. & Transfer & DV-Synth & DV-Human & Typed-dec. \\
\midrule
Qwen3.8-27B (dense) & 83.7 & 77.5 & 78.2 & 77.0 & 74.7 \\
Gemma-4-31B (dense) & 84.1 & 79.0 & 80.8 & 80.2 & 75.4 \\
Qwen3.5-397B-A17B (MoE) & 86.5 & 80.7 & 75.8 & 79.4 & 73.8 \\
\bottomrule
\end{tabular}
\end{table}

\section{Gemma with its own reasoning labels}
\label{app:gself}
In the controlled study, every base is trained on \sprd{}-C labels from Qwen3.8-27B. To test whether Gemma's smaller gains come from this cross-family teacher, we relabelled the same 16,040 pool items with Gemma-4-31B-it in thinking mode (four samples each; 0.65\% unparsed) and applied the same \sprd{}-C filter. Gemma's majority agrees with Qwen-27B's on 85.3\% of the pool, and the filter keeps 10,530 items (Qwen: 10,331; 9,558 in common), so these variants train about 2\% more steps with the same recipe. The items themselves are still posed by Qwen3.8-27B. These variants were trained after the pre-registration and are post hoc; the pre-registered estimates use the Qwen-label variants.

\begin{table}[h]
\centering\small
\caption{Gemma-4-31B-it, three-seed means (accuracy, \%), validation sets. Own labels: \sprd{}-C relabelled by Gemma's own reasoning.}
\label{tab:gself}
\begin{tabular}{lrrrr}
\toprule
Variant & Val. & DV-Dev-A & DV-Dev-B & Transfer \\
\midrule
Baseline, step-matched & 82.4 & 88.9 & 86.3 & 77.6 \\
+ \sprd{}-C, Qwen-27B labels & 83.2 & 91.1 & 89.6 & 78.0 \\
+ \sprd{}-C, own labels & 83.7 & 91.4 & 89.9 & 78.6 \\
+ \dds{} + \sprd{}-C, Qwen-27B labels & 84.1 & 91.1 & 90.7 & 79.0 \\
+ \dds{} + \sprd{}-C, own labels & 83.3 & 92.1 & 91.3 & 78.4 \\
\bottomrule
\end{tabular}
\end{table}

\begin{table}[h]
\centering\small
\caption{The same Gemma-4-31B-it variants on public and test sets (three-seed means, accuracy \%; ECE on the transfer suite before $\rightarrow$ after temperature scaling). All reads of the own-label variants are post hoc and descriptive; they are not part of the pre-registered contrasts.}
\label{tab:gselfsealed}
\setlength{\tabcolsep}{4pt}
\begin{tabular}{lrrrrr}
\toprule
Variant & Nimble-13 & Typed-dec. & DV-Synth & DV-Human & ECE \\
\midrule
Baseline, step-matched & 75.3 & 74.1 & 79.9 & 79.5 & 22.1$\rightarrow$5.4 \\
+ \sprd{}-C, Qwen-27B labels & 76.6 & 75.1 & 79.6 & 79.4 & 21.6$\rightarrow$6.0 \\
+ \sprd{}-C, own labels & 76.7 & 75.3 & 80.6 & 80.0 & 21.0$\rightarrow$5.9 \\
+ \dds{} + \sprd{}-C, Qwen-27B labels & 76.3 & 75.4 & 80.8 & 80.2 & 20.7$\rightarrow$5.4 \\
+ \dds{} + \sprd{}-C, own labels & 77.1 & 75.3 & 80.6 & 80.2 & 21.2$\rightarrow$6.1 \\
\bottomrule
\end{tabular}
\end{table}

Using Gemma's own reasoning labels instead of Qwen-27B's gives comparable results in these runs: every seed-paired difference lies within seed variation (full recipe, own minus Qwen labels: $+$0.8 on Nimble-13, $-$0.1 on typed-decisions, $-$0.3 on DV-Synth, $+$0.1 on DV-Human), and calibration is unchanged. The full recipe with own labels still improves on step-matched baseline on average on all four sets ($+$1.8, $+$1.2, $+$0.7, $+$0.8), with all three seeds positive on Nimble-13, typed-decisions and DV-Human. This control does not support teacher family as the sole explanation for Gemma's smaller gains; it changes both the retained subset and the training-step count and is not a formal equivalence test.

\section{Tuning as adaptation of the readout}
\label{app:adapt}
Table~\ref{tab:adapt} reports the size ladder from a scaling study with an earlier decision-tuning recipe (LoRA rank 16 on 43.9k records of public, Nimble and in-distribution data; one seed; Qwen3.5 dense bases), scored as the mean of the validation suite, DV-Dev-A/B and the eight choice and yes/no subsets of Nimble-13. The untuned single-pass readout already reaches 76\% of the tuned accuracy at 0.8B and 97\% at 27B; the tuning gain shrinks steadily with size; and 1/16 of the data (2.7k records) recovers most of the gain at every size. At 4B and above, scaling the learned update down to 0.65 after training helps (+0.3 to +0.8), consistent with tuning that adapts rather than replaces the pretrained representation.

\begin{table}[h]
\centering\small
\caption{Untuned readout versus decision tuning across Qwen3.5 sizes (accuracy, \%, mean of four development sets; scaling study, one seed). Share: untuned as a fraction of fully tuned accuracy. 1/16: tuned on 1/16 of the data.}
\label{tab:adapt}
\begin{tabular}{lrrrrrr}
\toprule
Base & Untuned & 1/16 data & Full data & Full, $s{=}0.65$ & Gain & Share \\
\midrule
0.8B & 50.9 & 60.8 & 66.7 & 65.8 & +15.8 & 76\% \\
2B & 59.9 & 68.3 & 73.0 & 72.0 & +13.1 & 82\% \\
4B & 72.1 & 76.9 & 79.9 & 80.2 & +7.8 & 90\% \\
9B & 76.0 & 81.2 & 81.9 & 82.6 & +5.9 & 93\% \\
27B & 84.3 & -- & 86.6 & 87.4 & +2.3 & 97\% \\
\bottomrule
\end{tabular}
\end{table}

\section{Is a classification head enough?}
\label{app:head}
The readout of Eq.~\ref{eq:readout} is already a linear head: the last hidden state multiplied by the output-layer rows of the option letters. We trained only these 26 rows (106k parameters, against about 40M for LoRA rank 16) of Qwen3.5-9B, whose output rows are untied from the input embeddings, with the backbone frozen and no bias, so the trained head drops into the unmodified scorer. Training data are identical to the baseline and \dds{} variants, with four option-order views per item. Learning rate, weight decay and epochs were selected on the validation suite from 120 settings, whereas LoRA used its fixed recipe, which favours the head. Three seeds; the test sets were read post hoc and these variants are descriptive.

\begin{table}[h]
\centering\small
\caption{Head-only training versus LoRA on Qwen3.5-9B (three-seed means, accuracy \%; ECE on the transfer suite before $\rightarrow$ after temperature scaling).}
\label{tab:head}
\setlength{\tabcolsep}{4pt}
\begin{tabular}{lrrrrrrr}
\toprule
Variant & Val. & Transfer & Nimble-13 & Typed-dec. & DV-Synth & DV-Human & ECE \\
\midrule
Untuned & 71.8 & 66.1 & 73.7 & 63.5 & 52.5 & 69.4 & 17.1$\rightarrow$6.5 \\
Head only, baseline & 72.7 & 66.2 & 73.4 & 64.2 & 52.7 & 69.5 & 3.1$\rightarrow$6.2 \\
LoRA, baseline & 75.9 & 69.7 & 74.1 & 69.8 & 64.5 & 72.8 & 23.8$\rightarrow$4.3 \\
Head only, \dds{} mix & 72.6 & 66.2 & 73.8 & 64.9 & 52.2 & 69.5 & 4.3$\rightarrow$6.2 \\
LoRA, \dds{} mix & 77.1 & 71.6 & 71.6 & 72.2 & 66.1 & 75.0 & 10.0$\rightarrow$2.7 \\
\bottomrule
\end{tabular}
\end{table}

Training only the head recovers little of the fine-tuning gain. On the \dds{} mixture it changes the transfer suite by $+$0.1 (LoRA $+$5.6), typed-decisions by $+$1.3 ($+$8.7), DV-Synth by $-$0.2 ($+$13.6) and DV-Human by $+$0.1 ($+$5.6), computed from unrounded means. Validation accuracy plateaus near 72.8 across all 120 settings, so the frozen features, not the head, limit transfer. The head mainly corrects the logit scale: raw ECE on the transfer suite falls from 17.1 to 3--4, but after temperature scaling it is no better than the untuned model (6.2 vs.\ 6.5). Decision supervision therefore has to adapt the backbone, and the \dds{} mixture helps only when it does. LoRA's drop on Nimble-13 with the \dds{} mix comes from the SummEval subsets' seed instability (Table~\ref{tab:public13}); the head is flat there. Caveats: one base, and a linear letter-row head rather than an MLP.

\section{Further analyses}
\label{app:moved}
\paragraph{Where reasoning helps and where it hurts.}
Table~\ref{tab:devgap} (Appendix~\ref{app:extra}) breaks the untuned model's rollout--readout gap down by source in the validation suite.
Reasoning helps most on WinoGrande (+26.0), MedMCQA (+7.5) and irony (+6.5), is neutral on knowledge-bound ARC and CaseHOLD, and hurts on CommitmentBank ($-16.1$).
The CommitmentBank errors are systematic: in 11 of the 15 items that reasoning gets wrong and the readout gets right, reasoning moves a gold \emph{entailment} or \emph{contradiction} to \emph{neutral} or \emph{not determined}.
CommitmentBank labels speaker commitment under embedding predicates, which relies on pragmatic projection; deliberate reasoning tends to refuse such inferences as unsupported.
This is a bias \sprd{} could transfer to the student. It does not: averaged over three seeds, the 27B \sprd{} student is within one item of its baseline counterpart on CommitmentBank ($-1.2$ points on 56 items), while its largest validation gains are on the two datasets where the teacher's gap was largest, WinoGrande (+2.8) and irony (+3.2).

\paragraph{The same check on a second model family.}
On Gemma-4-31B (recipe with benchmark-source data, three \sprd{}-C seeds against one step-matched baseline seed), restricting DV-Dev-A and DV-Dev-B to items that Magistral also confirms leaves a gain of +2.0 points (family-level 95\% CI [0.3, 3.9]); on this model family the gain does not depend on contested labels.

\paragraph{Cost details.}
The \sprd{} teacher pass over 16k pool items generated about 59M tokens (about 8 H100-hours with vLLM in FP8); \dds{} synthesis is API-based and needs no reasoning pass.
Both leave inference unchanged: one prefill and no generated tokens, against a median of 335--513 generated tokens per sample for the reasoning rollout.
\begin{table}[t]
\centering\small
\caption{Calibration on the human-labelled transfer suite (seed means). TS: temperature scaling fitted on the validation suite by NLL. Temperature scaling removes most of the ECE gap to step-matched baseline; the recipes keep a lower Brier score on every base. Kev, which ships a calibrated temperature, has the lowest raw ECE but higher NLL and Brier than our models. Gemma's fitted temperatures are 10--12 for every variant: its raw readout is far more overconfident than Qwen's (1.1--3.4).}
\label{tab:calib}
\setlength{\tabcolsep}{3pt}
\fitbox{%
\begin{tabular}{llrrrrrr}
\toprule
 & & \multicolumn{2}{c}{ECE (\%)} & \multicolumn{2}{c}{NLL} & \multicolumn{2}{c}{Brier} \\
\cmidrule(lr){3-4}\cmidrule(lr){5-6}\cmidrule(lr){7-8}
Base & Recipe & raw & TS & raw & TS & raw & TS \\
\midrule
9B & Untuned & 17.0 & 6.4 & 1.20 & 0.94 & 0.501 & 0.460 \\
 & Nimble-9B & 15.0 & \textbf{2.5} & 1.16 & 0.89 & 0.474 & 0.436 \\
 & Kev-9B & \textbf{3.7} & \textbf{2.4} & 0.95 & 0.93 & 0.466 & 0.464 \\
 & Baseline, step-m. & 23.8 & 4.3 & 1.50 & 0.84 & 0.527 & 0.420 \\
\rowcolor{gray!15}  & +\dds{} & 10.0 & 2.7 & 0.91 & 0.80 & 0.407 & 0.390 \\
\rowcolor{gray!15}  & +\sprd{}-C & 11.7 & 3.4 & 0.93 & 0.80 & 0.417 & 0.393 \\
\rowcolor{gray!15}  & +both & 9.7 & 3.0 & \textbf{0.89} & \textbf{0.79} & \textbf{0.404} & \textbf{0.388} \\
\midrule
27B & Untuned & 6.2 & 4.6 & 0.85 & 0.82 & 0.359 & 0.358 \\
 & OpenJev-27B & 6.9 & 6.9 & 0.80 & 0.80 & 0.348 & 0.348 \\
 & Kev-27B & \textbf{3.0} & \textbf{3.0} & 0.81 & 0.82 & 0.370 & 0.371 \\
 & Baseline, step-m. & 18.5 & 5.7 & 1.15 & \textbf{0.71} & 0.415 & 0.344 \\
\rowcolor{gray!15}  & +\dds{} & 8.3 & 3.9 & 0.81 & 0.73 & 0.344 & 0.334 \\
\rowcolor{gray!15}  & +\sprd{}-C & 11.0 & 4.4 & 0.84 & \textbf{0.71} & 0.352 & 0.332 \\
\rowcolor{gray!15}  & +both & 8.9 & 4.5 & \textbf{0.80} & 0.72 & \textbf{0.338} & \textbf{0.328} \\
\midrule
31B & Untuned & 22.0 & \textbf{5.4} & 5.61 & 0.78 & 0.442 & 0.356 \\
 & Baseline, step-m. & 22.1 & \textbf{5.4} & 5.28 & 0.76 & 0.443 & 0.350 \\
\rowcolor{gray!15}  & +\dds{} & 20.8 & \textbf{5.4} & \textbf{4.70} & \textbf{0.71} & 0.418 & \textbf{0.326} \\
\rowcolor{gray!15}  & +\sprd{}-C & 21.6 & 6.0 & 4.92 & 0.73 & 0.434 & 0.337 \\
\rowcolor{gray!15}  & +both & \textbf{20.7} & \textbf{5.4} & 4.74 & 0.72 & \textbf{0.416} & 0.328 \\
\bottomrule
\end{tabular}}
\end{table}

\begin{table}[t]
\centering\small
\caption{Calibration on the test sets after temperature scaling with each variant's temperature fitted on the validation suite (seed means; post hoc). The fitted temperature transfers (ECE 2.5--7 for every variant); every recipe has lower NLL and Brier score than the step-matched baseline on every base.}
\label{tab:sealedcalib}
\setlength{\tabcolsep}{3pt}
\fitbox{%
\begin{tabular}{llrrrrrr}
\toprule
 & & \multicolumn{3}{c}{DV-Synth} & \multicolumn{3}{c}{DV-Human} \\
\cmidrule(lr){3-5}\cmidrule(lr){6-8}
Base & Recipe & ECE & NLL & Brier & ECE & NLL & Brier \\
\midrule
9B & Baseline, step-m. & 6.7 & 0.86 & 0.464 & 4.5 & 0.67 & 0.363 \\
\rowcolor{gray!15}  & +\dds{} & 5.7 & 0.81 & 0.436 & 3.6 & 0.61 & 0.329 \\
\rowcolor{gray!15}  & +\sprd{}-C & \textbf{4.2} & \textbf{0.77} & \textbf{0.418} & \textbf{2.7} & \textbf{0.61} & 0.327 \\
\rowcolor{gray!15}  & +both & 6.1 & \textbf{0.78} & 0.423 & 3.3 & \textbf{0.60} & \textbf{0.325} \\
\midrule
27B & Baseline, step-m. & 5.1 & 0.55 & 0.296 & 4.1 & 0.63 & 0.336 \\
\rowcolor{gray!15}  & +\dds{} & \textbf{3.0} & 0.49 & 0.269 & 2.6 & 0.57 & 0.306 \\
\rowcolor{gray!15}  & +\sprd{}-C & 4.2 & 0.49 & 0.267 & 2.8 & 0.57 & 0.305 \\
\rowcolor{gray!15}  & +both & 4.0 & \textbf{0.46} & \textbf{0.258} & \textbf{2.3} & \textbf{0.56} & \textbf{0.301} \\
\midrule
31B & Baseline, step-m. & 6.0 & 0.62 & 0.307 & 4.6 & 0.59 & 0.301 \\
\rowcolor{gray!15}  & +\dds{} & 6.9 & \textbf{0.55} & \textbf{0.273} & 4.1 & 0.57 & 0.292 \\
\rowcolor{gray!15}  & +\sprd{}-C & 7.1 & 0.59 & 0.291 & 5.1 & 0.58 & 0.294 \\
\rowcolor{gray!15}  & +both & \textbf{5.9} & 0.56 & 0.279 & \textbf{4.0} & \textbf{0.56} & \textbf{0.284} \\
\bottomrule
\end{tabular}}
\end{table}

\paragraph{DecisionVista-Human by source.}
\begin{table}[h]
\centering\small
\caption{DecisionVista-Human gain of \dds{}+\sprd{}-C over the step-matched baseline by source (points), over the seven pre-registered seed pairs (three at 9B, two on the 27B and 31B bases). Clusters: distinct documents or premise sets; a cluster bootstrap over them gives the same pooled interval as the item bootstrap ([0.9, 2.2]).}
\label{tab:hssource}
\setlength{\tabcolsep}{3pt}
\begin{tabular}{lrrrr}
\toprule
Source & Clust. & Mean $\Delta$ & Seed-pair range & $>0$ \\
\midrule
ContractNLI & 242 & +2.7 & [+0.7, +7.1] & 7/7 \\
CSQA2 & 300 & +3.1 & [$-$0.3, +5.7] & 6/7 \\
ETHICS & 300 & $-$1.3 & [$-$3.7, +1.3] & 2/7 \\
Fig-QA & 300 & +0.9 & [$-$0.7, +2.3] & 6/7 \\
FOLIO & 73 & +1.5 & [$-$2.0, +6.4] & 3/7 \\
LIAR & 299 & +0.4 & [$-$4.7, +6.4] & 3/7 \\
SciCite & 300 & +4.7 & [$-$0.7, +9.0] & 6/7 \\
SciFact & 251 & +1.0 & [$-$1.7, +3.7] & 5/7 \\
\bottomrule
\end{tabular}
\end{table}

\begin{table}[t]
\centering\small
\caption{Same items, same steps, different labels (Qwen3.8-27B as teacher; baseline data plus the 10,331 \sprd{}-C items; two seeds each; no \dds{}). On this subset the hard reasoning label equals the generator's intended label by construction.}
\label{tab:samesubset}
\setlength{\tabcolsep}{4pt}
\fitbox{%
\begin{tabular}{llrrrr}
\toprule
 & Labels on the 10.3k items & Val. & DV-Dev-A & DV-Dev-B & DV-Pilot \\
\midrule
\multirow{3}{*}{27B} & \cellcolor{gray!15}reasoning, soft (\sprd{}-C) & \cellcolor{gray!15}\textbf{83.5} & \cellcolor{gray!15}\textbf{90.3} & \cellcolor{gray!15}89.1 & \cellcolor{gray!15}98.9 \\
 & reasoning, hard & 83.1 & 89.9 & \textbf{89.2} & \textbf{99.0} \\
 & own single-pass readout & 81.9 & 89.0 & 86.7 & 98.9 \\
\midrule
\multirow{3}{*}{9B} & \cellcolor{gray!15}reasoning, soft (\sprd{}-C) & \cellcolor{gray!15}76.6 & \cellcolor{gray!15}87.1 & \cellcolor{gray!15}82.7 & \cellcolor{gray!15}\textbf{96.6} \\
 & reasoning, hard & \textbf{77.4} & \textbf{87.2} & \textbf{83.4} & 96.2 \\
 & 27B single-pass readout & \textbf{77.4} & 86.9 & 81.1 & 96.1 \\
\bottomrule
\end{tabular}}
\end{table}

\paragraph{Label validity on the earlier sets (details).}
DV-Dev-A, DV-Dev-B and DV-Stress keep an item when gpt-5.5 (author) and gpt-6-astra (blind) agree.
To check these labels independently we had Magistral-Small, a third model family used nowhere else in the pipeline, answer every item blind (majority of three reasoning samples).
It agrees with the reference on 90.2\%, 87.8\% and 87.4\% of items; on DV-Stress agreement ranges from 100\% on some families to 8\% on \emph{contract renewal} (12 items, where every model we trained scores below 30\% and the untuned model 42\%, suggesting faulty references).
Restricting the 27B comparison to items where all three families agree shrinks \sprd{}'s gain over baseline from +1.0 to +0.4 on DV-Stress and from +1.9 to +0.8 across all 48 families (CI [$-$0.3, 2.0]).
On the 279 contested items, \sprd{} answers with the reference label 71\% of the time and baseline tuning 61\%: distilling reasoning moves the student toward the answers that the reasoning models used to build the benchmark would give.
Whether that is a gain in correctness or in agreement with a shared convention cannot be settled without human judgements of these items.
It is the reason we built DV-Pilot from three-way agreement, and we suggest that work on distilling reasoning report results on independently validated subsets.

\paragraph{Test-time option order.}
All scores use the canonical option order. Re-scoring DecisionVista-Synth and DecisionVista-Human with options permuted per item (seed 1) for step-matched baseline and the combined recipe (seed 17 on each base; post hoc) changes accuracy by at most 1.4 points and leaves 89--97\% of predictions unchanged. The recipe's advantage keeps its sign in five of six cases (DecisionVista-Synth: 9B +2.7$\to$+2.5, 27B +2.1$\to$+2.0, 31B +0.6$\to$+1.4; DecisionVista-Human: 9B +2.0$\to$+1.9, 27B +2.3$\to$+1.9, 31B +0.5$\to$0.0).

\begin{table}[t]
\centering\small
\caption{Individual components trained for the full recipe's 4,031 steps (post hoc), against baseline trained for 4,032 steps. Qwen3.5-9B: seeds 17/29/41; Qwen3.8-27B: seeds 17/29. Transfer: gain in item accuracy on the transfer suite (no interval).}
\label{tab:stepcomp}
\setlength{\tabcolsep}{3pt}
\fitbox{%
\begin{tabular}{lrccc}
\toprule
vs.\ step-matched baseline & Transfer & DV-Synth & DV-Human & Typed-dec. \\
\midrule
\multicolumn{5}{l}{\emph{Qwen3.5-9B}} \\
+\dds{}, one pass & +1.9 & \textbf{+1.6} {\scriptsize[0.07, 3.2]} & \textbf{+2.2} {\scriptsize[1.2, 3.2]} & \textbf{+2.4} {\scriptsize[1.0, 3.8]} \\
+\dds{}, 4,031 steps & +2.2 & +1.2 {\scriptsize[$-$1.1, 3.3]} & \textbf{+1.8} {\scriptsize[0.8, 2.9]} & +0.0 {\scriptsize[$-$1.2, 1.3]} \\
+\sprd{}-C, one pass & +1.4 & \textbf{+2.1} {\scriptsize[0.02, 4.0]} & \textbf{+2.1} {\scriptsize[1.1, 3.1]} & +1.3 {\scriptsize[$-$0.2, 2.8]} \\
+\sprd{}-C, 4,031 steps & +1.6 & \textbf{+2.6} {\scriptsize[0.8, 4.4]} & \textbf{+1.9} {\scriptsize[1.0, 2.9]} & \textbf{+1.9} {\scriptsize[0.3, 3.4]} \\
\rowcolor{gray!15} +both, 4,031 steps & +2.2 & \textbf{+2.3} {\scriptsize[0.6, 3.9]} & \textbf{+2.1} {\scriptsize[1.1, 3.1]} & \textbf{+2.5} {\scriptsize[0.9, 4.2]} \\
\multicolumn{5}{l}{\emph{Qwen3.8-27B}} \\
+\dds{}, one pass & +1.1 & +1.2 {\scriptsize[$-$0.9, 3.2]} & +1.0 {\scriptsize[$-$0.04, 2.1]} & \textbf{+2.7} {\scriptsize[1.4, 3.9]} \\
+\dds{}, 4,031 steps & +1.3 & +2.5 {\scriptsize[$-$0.01, 5.0]} & \textbf{+1.6} {\scriptsize[0.6, 2.5]} & +1.2 {\scriptsize[$-$0.1, 2.4]} \\
+\sprd{}-C, one pass & +0.4 & +1.6 {\scriptsize[$-$1.0, 4.6]} & +0.9 {\scriptsize[$-$0.05, 1.9]} & \textbf{+3.1} {\scriptsize[1.7, 4.6]} \\
+\sprd{}-C, 4,031 steps & +0.8 & \textbf{+2.9} {\scriptsize[1.2, 4.7]} & \textbf{+1.4} {\scriptsize[0.5, 2.3]} & \textbf{+3.4} {\scriptsize[2.0, 4.8]} \\
\rowcolor{gray!15} +both, 4,031 steps & +1.4 & +2.4 {\scriptsize[$-$0.3, 5.3]} & \textbf{+1.9} {\scriptsize[0.8, 3.0]} & \textbf{+2.9} {\scriptsize[1.5, 4.2]} \\
\bottomrule
\end{tabular}}
\end{table}

\begin{table}[t]
\centering\small
\caption{Post hoc: the combined recipe against a baseline selected on the validation suite. Per seed, the baseline is selected from one-pass and step-matched training, each with the full or a 0.65-scaled update; the recipe gets the same selection over its update scale. Seeds as pre-registered; same bootstraps as Table~\ref{tab:contrast}.}
\label{tab:devsel}
\setlength{\tabcolsep}{3pt}
\fitbox{%
\begin{tabular}{lccc}
\toprule
vs.\ validation-selected baseline & DV-Synth & DV-Human & Typed-dec. \\
\midrule
Qwen3.5-9B & \textbf{+1.6} {\scriptsize[0.3, 2.9]} & +0.6 {\scriptsize[$-$0.2, 1.5]} & \textbf{+1.4} {\scriptsize[0.2, 2.6]} \\
Qwen3.8-27B & +0.7 {\scriptsize[$-$1.3, 2.6]} & \textbf{+1.4} {\scriptsize[0.1, 2.8]} & \textbf{+2.6} {\scriptsize[1.1, 4.1]} \\
Gemma-4-31B & +0.1 {\scriptsize[$-$1.2, 1.4]} & +0.3 {\scriptsize[$-$0.6, 1.1]} & \textbf{+2.2} {\scriptsize[0.6, 3.8]} \\
\midrule Mean across bases & +0.8 {\scriptsize[$-$0.1, 1.7]} & \textbf{+0.8} {\scriptsize[0.2, 1.4]} & \textbf{+2.1} {\scriptsize[1.2, 3.0]} \\
\bottomrule
\end{tabular}}
\end{table}

\begin{table}[t]
\centering\small
\caption{Label-source controls from the \sprd{} study (27B; baseline data plus the self-posed pool labelled by each source; same trainer and steps within the pool rows; two seeds), read post hoc on the three sets no design decision used. The bottom rows train on the identical 10,331 \sprd{}-C items.}
\label{tab:labelctl}
\setlength{\tabcolsep}{3pt}
\fitbox{%
\begin{tabular}{lrrr}
\toprule
Qwen3.8-27B, labels & DV-Synth & DV-Human & Typed-dec. \\
\midrule
Step-matched gold (no pool) & 77.1 & 76.6 & 72.2 \\
\midrule
Pool, own readout labels & 76.1 & 76.2 & 73.0 \\
Pool, generator's intended labels & 78.0 & 77.2 & 72.4 \\
Pool, gpt-5.5 labels & \textbf{78.7} & \textbf{77.6} & 74.1 \\
\rowcolor{gray!15} Pool, reasoning labels (\sprd{}) & 77.7 & 76.5 & 75.0 \\
\midrule
\rowcolor{gray!15} \sprd{}-C subset, reasoning labels & 76.7 & 76.9 & \textbf{75.1} \\
\sprd{}-C subset, own readout labels & 75.0 & 75.7 & 73.2 \\
\bottomrule
\end{tabular}}
\end{table}

\paragraph{Label validity on the earlier sets (summary).}
On DV-Dev-A, DV-Dev-B and DV-Stress, where labels come from two models, a third family (Magistral) disputes 10--13\% of items, and at 27B most of \sprd{}'s gain over baseline sits on those contested items (+1.9 over all 48 families, +0.8 [$-$0.3, 2.0] on items all three families confirm; Appendix~\ref{app:moved}). Whether this is correctness or agreement with a shared model convention needs human judgement; it is why DV-Pilot and DecisionVista-Synth require multi-family agreement and why DecisionVista-Human is human-labelled.

\paragraph{What in \dds{} helps.} Exploratory single-seed 4B ablations (Appendix Table~\ref{tab:ddsabl}) suggest that a few thousand audited synthesised decision tasks give most of the gain and that removing any one field changes little; we test \dds{} only as a package.

\paragraph{Transfer sensitivity analyses.}
Both primary intervals remain above zero when training seeds are resampled together with families or items (DecisionVista-Synth [0.2, 3.5], DecisionVista-Human [0.8, 2.3]), when whole decision operators (DecisionVista-Synth [0.5, 3.1]) or documents (DecisionVista-Human [0.9, 2.2]) are resampled, without the one DecisionVista-Synth author with a role in training data (+1.9 [0.6, 3.3] on the 25 families gpt-6-astra did not write; Table~\ref{tab:secondary}), while dropping any one DecisionVista-Human source retains a positive mean gain (+1.1 to +2.0).
Adding seed-41 runs trained after pre-registration leaves both point estimates unchanged (Table~\ref{tab:secondary}).
The original exploratory single-component gains are 1.5--3.4 points, while the combination recovers 62--67\% of their sum.
Baseline tuning itself can improve new-schema accuracy at 27B on DecisionVista-Synth (Table~\ref{tab:fresh5}); on the validation sets, Gemma's step-matched baseline instead falls below its untuned base (84.3 vs.\ 85.8).
\paragraph{Calibration details.}
Kev's shipped temperature yields raw transfer-suite ECE of 2.4--3.7, while Gemma requires fitted temperatures of 10--12 across variants (Table~\ref{tab:calib}).

\section{Extended related work}
\label{app:related}

\paragraph{Decision interfaces and schema transfer.}
BERT classifiers commonly use task-specific output heads~\citep{devlin2019bert}.
TARS~\citep{halder2020tars} conditions on label descriptions and supports unseen classes, illustrating that encoder architectures can also support transfer.
Reading class decisions from next-token scores goes back to cloze-style prompting~\citep{schick2021pet} and is sensitive to label priors and wording~\citep{zhao2021calibrate}.
Decision services and models~\citep{jev,openjev,nimble,laya,kev} standardise this interface for typed questions; we name the class, specify what it should be evaluated on, and study how to train it for unseen schemas.

\paragraph{Schema diversity and synthetic task construction.}
Multi-task instruction tuning improves zero-shot transfer to held-out task types~\citep{wei2022flan,sanh2022t0,wang2022supernatural}; mixture design and data diversity affect the resulting capabilities~\citep{longpre2023flan,zhou2023lima}.
Self-Instruct and Unnatural Instructions use models to generate instruction tasks~\citep{wang2023selfinstruct,honovich2023unnatural}.
\dds{} combines synthetic variation of decision rules and label meanings with public and instruction coverage, evaluating the package against step-matched controls.
\sprd{} likewise uses self-posed schemas, audited against evaluation families, to obtain inputs for reasoning supervision.

\paragraph{Reasoning supervision and controls.}
Chain-of-thought~\citep{wei2022cot,kojima2022zeroshot} and self-consistency~\citep{wang2023selfconsistency} buy accuracy with generated tokens.
Rationale-based training transfers reasoning through generated explanations or intermediate targets~\citep{ho2023teachers,magister2023teaching,hsieh2023stepbystep,zelikman2022star,huang2023selfimprove}; it need not require rationale generation at deployment, as Distilling Step-by-Step permits label prediction alone~\citep{hsieh2023stepbystep}.
Implicit chain-of-thought~\citep{deng2023implicit,deng2024explicit} and System-2 distillation~\citep{yu2024distilling} also avoid explicit reasoning at inference, the latter by fine-tuning on self-consistent System-2 outputs.
\sprd{} is closest to System-2 distillation but supervises a one-position readout over run-time label sets with final-answer vote distributions, without student rationale training.
On-policy distillation trains on student-generated trajectories with teacher feedback~\citep{agarwal2024onpolicy}; on-policy self-distillation lets the same model teach with privileged context~\citep{zhao2026opsd}.
The shared idea is that stronger teacher behaviour can supervise a less capable computation mode.
\sprd{} instead uses a fixed pool and an offline frozen teacher, with no on-policy trajectories or constraint on distance from the base parameters.
In the primary study, the shared teacher is the student's own base only at 27B; Gemma's own-label runs are a separate post-hoc control.
Self-training~\citep{lee2013pseudo,xie2020noisy} provides our readout-label control, and weight-space interpolation~\citep{wortsman2022wiseft} our update-scaling baseline.

\paragraph{Routing and calibration.}
Selective classification abstains when confidence is low~\citep{geifman2017selective}; LLM cascades and routers send queries to a more expensive model when a cheaper one is unlikely to suffice~\citep{chen2023frugalgpt,ong2024routellm}.
Our router uses the student's untuned base using reasoning rollout as its fallback and a margin threshold fixed on the validation suite, with cost counted in generated tokens.
Temperature scaling~\citep{guo2017calibration} separately calibrates the readout probabilities.

\section{Serving efficiency measurement details}\label{app:speed}
Table~\ref{tab:speed} compares Gemma-4-31B serving efficiency on a validation/transfer workload.
The tuned readout uses a merged DDS + SPRD-C checkpoint trained with Gemma's own reasoning labels (G31X\_ldm\_gself).
Both timed models use BF16 vLLM serving on two H100 80GB GPUs each, with prefix caching disabled.
On 32 sequential validation/transfer inputs repeated twice, tuned readout has a median response latency of 40.6\,ms, versus 7.38\,s for reasoning rollout.
On 128 inputs at concurrency 32, also repeated twice, throughput is 44.2 and 0.787 requests/s, respectively.
Readout therefore reduces median latency by 182$\times$ and increases throughput by 56.2$\times$ relative to reasoning rollout.
The timer covers host-local HTTP requests through response parsing, excluding prompt preparation, model loading and compilation.
The serving API returns one option token to expose candidate scores, with no subsequent autoregressive decode step, while reasoning averages 1,027 output tokens in the concurrent workload.
Allocated GPU-hours per million requests are extrapolated as two GPUs times measured job duration divided by completed requests, scaled to one million; they exclude loading, compilation and training. Speedups in Table~\ref{tab:speed} are computed from the displayed median latencies.
These measurements do not measure confidence-router latency. The serving backend also applies Gemma's logit softcapping, unlike the original raw-logit scorer, so its probabilities require separate calibration before routing.

\end{document}